\documentclass[acmtog,screen,nonacm]{acmart}

\usepackage{booktabs} 

\usepackage[ruled]{algorithm2e} 

\SetAlFnt{\small}
\SetAlCapFnt{\small}
\SetAlCapNameFnt{\small}
\SetAlCapHSkip{0pt}

\usepackage{amsmath}

\usepackage{subcaption}

\usepackage{tkz-euclide}
\usepackage{tikz}
\usepackage{pgfplots}
\pgfplotsset{compat=1.14}
\usetikzlibrary{positioning,calc,fadings,backgrounds,fit,3d,shapes.misc,shapes.geometric}
\usetikzlibrary{matrix}

\usepackage{multirow}

\usepackage{overpic}
\usepackage{color, colortbl}
\usepackage{xcolor}
\definecolor{sim}{RGB}{128,0,128}
\definecolor{interp}{RGB}{0, 255, 0}

\usepackage[capitalize]{cleveref}
\crefname{section}{Sec.}{Secs.}
\Crefname{section}{Section}{Sections}
\Crefname{table}{Table}{Tables}
\crefname{table}{Tab.}{Tabs.}

\makeatletter
\DeclareRobustCommand\onedot{\futurelet\@let@token\@onedot}
\def\@onedot{\ifx\@let@token.\else.\null\fi\xspace}

\def\eg{\emph{e.g}\onedot} 
\def\ie{\emph{i.e}\onedot}

\makeatother

\definecolor{mytbcol}{RGB}{175,227,246}

\usepackage{threeparttable}

\renewcommand\footnotetextcopyrightpermission[1]{} 
\begin{document}
\title{3DShape2VecSet: A 3D Shape Representation for Neural Fields and Generative Diffusion Models}

\author[]{Biao Zhang}
\affiliation{
    \institution{KAUST}
    \country{Saudi Arabia}
}
\email{biao.zhang@kaust.edu.sa}

\author[]{Jiapeng Tang}
\affiliation{
    \institution{TU Munich}
    \country{Germany}
}
\email{jiapeng.tang@tum.de}

\author[]{Matthias Nie{\ss}ner}
\affiliation{
    \institution{TU Munich}
    \country{Germany}
}
\email{niessner@tum.de}

\author[]{Peter Wonka}
\affiliation{
    \institution{KAUST}
    \country{Saudi Arabia}
}
\email{peter.wonka@kaust.edu.sa}

\begin{abstract}
We introduce 3DShape2VecSet, a novel shape representation for neural fields designed for generative diffusion models.
Our shape representation can encode 3D shapes given as surface models or point clouds, and represents them as neural fields.
The concept of neural fields has previously been combined with a global latent vector, a regular grid of latent vectors, or an irregular grid of latent vectors.
Our new representation encodes neural fields on top of a set of vectors.
We draw from multiple concepts, such as the radial basis function representation and the cross attention and self-attention function, to design a learnable representation that is especially suitable for processing with transformers.
Our results show improved performance in 3D shape encoding and 3D shape generative modeling tasks. We demonstrate a wide variety of generative applications: unconditioned generation, category-conditioned generation, text-conditioned generation, point-cloud completion, and image-conditioned generation.
Code: \url{https://1zb.github.io/3DShape2VecSet/}.
\end{abstract}

%
%




\begin{teaserfigure}
    \centering
    \begin{overpic}[trim={0cm 0cm 0cm 0cm},clip,width=1\linewidth,grid=false]{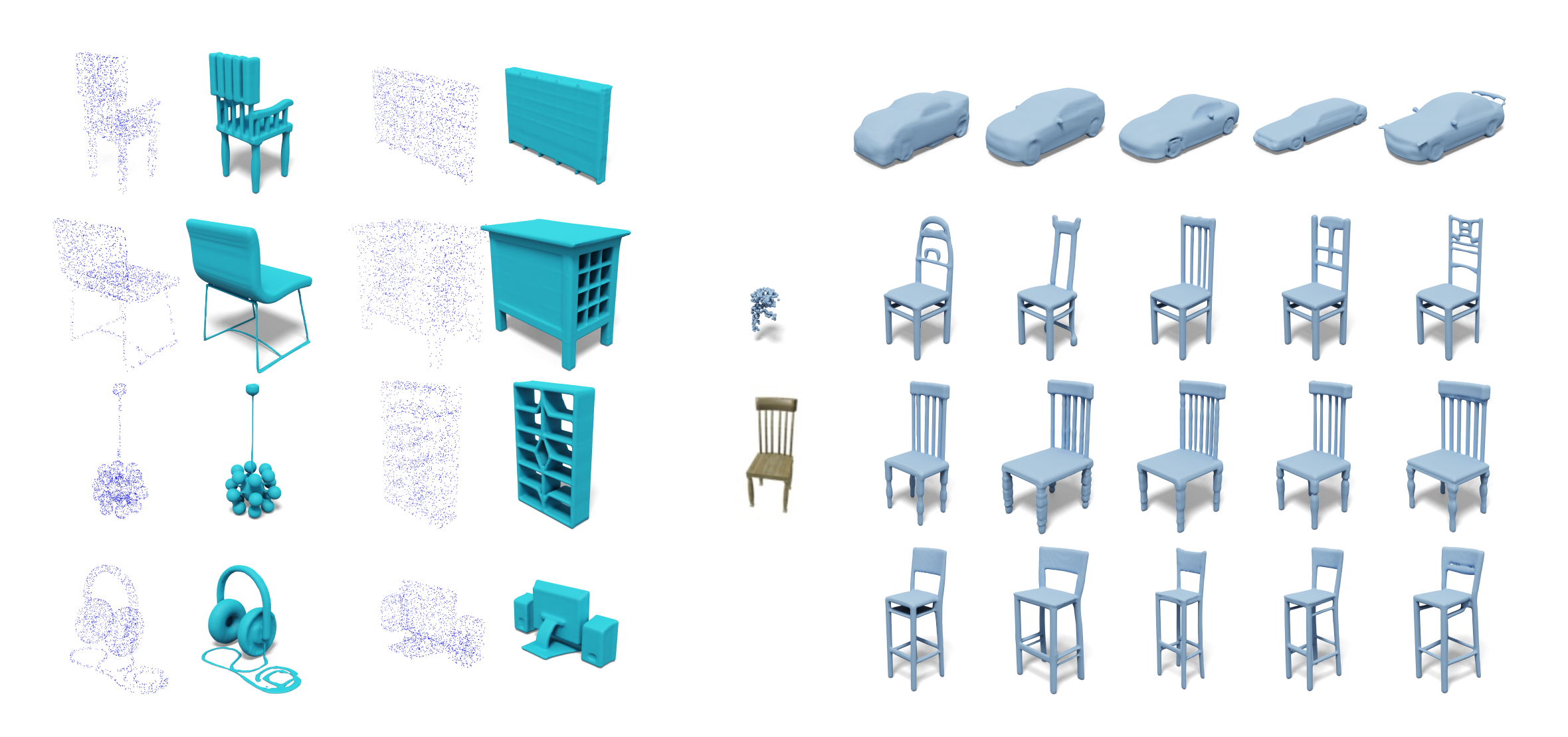}
        \put(5, 45){\small{Input}}
        \put(11, 45){\small{Reconstruction}}
        \put(25, 45){\small{Input}}
        \put(31, 45){\small{Reconstruction}}
        \dashline{0.7}(43,2)(43,46)
        \put(46, 45){\small{Condition}}
        \put(72, 45){\small{Generation}}
        \put(44.5, 6){\small{``the tallest chair''}}
        \put(47, 39){\small{car}}
    \end{overpic}
    \vspace{-20pt}
    \caption{\textbf{Left:} Shape autoencoding results (surface reconstruction from point clouds) \textbf{Right:} the various down-stream applications of \textbf{3DShape2VecSet} (from top to down): (a) category-conditioned generation; (b) point clouds conditioned generation (shape completion from partial point clouds); (c) image conditioned generation (shape reconstruction from single-view images); (d) text-conditioned generation.}
    \label{fig:teaser}
\end{teaserfigure}
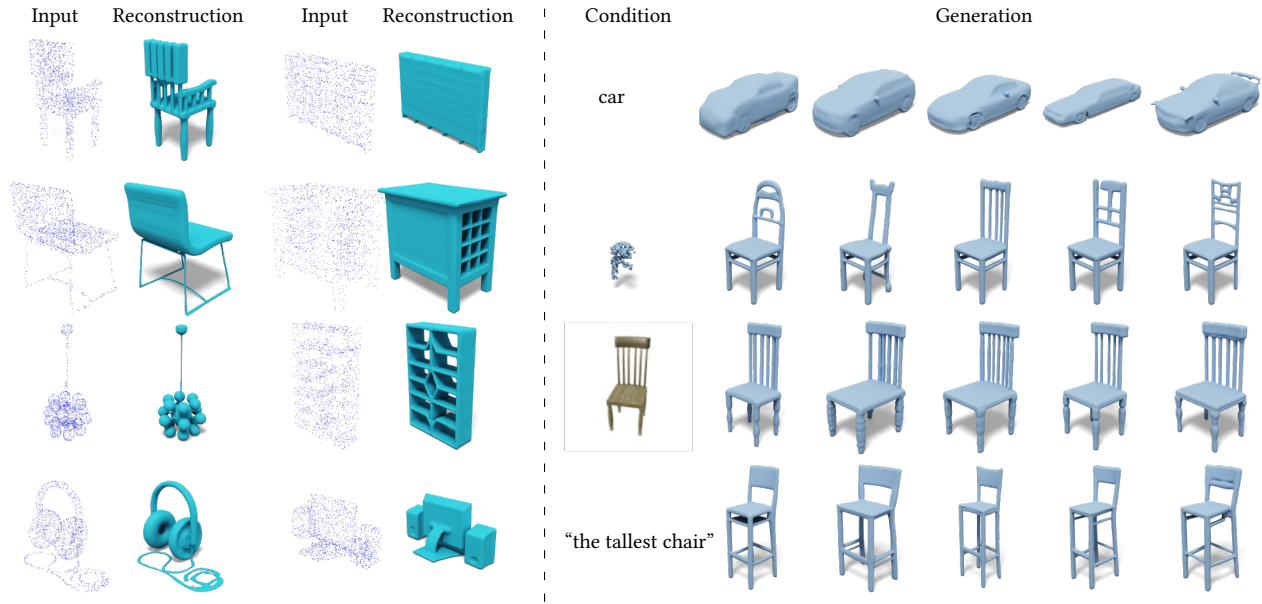

\maketitle

\section{Introduction}

The ability to generate realistic and diverse 3D content has many potential applications, including computer graphics, gaming, and virtual reality. 
To this end, many generative models have been explored, \eg, generative adversarial networks, variational autoencoders, normalizing flows, and autoregressive models. Recently, diffusion models have emerged as one of the most popular method with fantastic results in the 2D image domain~\cite{ho2020denoising,rombach2022high} and have shown their superiority over other generative methods. For instance, it is possible to do unconditional generation~\cite{rombach2022high,Karras2022edm}, text conditioned generation~\cite{saharia2022photorealistic,rombach2022high}, and generative image inpainting~\cite{Lugmayr2022RePaintIU}. However, the success in the 2D domain has not yet been matched in the 3D domain.

In this work, we will study diffusion models for 3D shape generation. One major challenge in adapting 2D diffusion models to 3D is the design of a suitable shape representation. The design of such a shape representation is the major focus of our work, and we will discuss several design choices that lead to the development of our proposed representation.

Different from 2D images, there are several predominant ways to represent 3D data, \eg, voxels, point clouds, meshes, and neural fields. In general, we believe that surface-based representations are more suitable for downstream applications than point clouds. Among the available choices, we choose to build on neural fields as they have many advantages: they are continuous, represent complete surfaces and not only point samples, and they enable many interesting combinations of traditional data structure design and representation learning using neural networks. 


Two major approaches for 2D diffusion models are to either use a compressed latent space, \eg, latent diffusion~\cite{rombach2022high}, or to use a sequence of diffusion models of increasing resolution, \eg,~\cite{saharia2022photorealistic,ramesh2022}. While both of these approaches seem viable in 3D, our initial experiments indicated that it is much easier to work with a compressed latent space. We therefore follow the latent diffusion approach.

A subsequent design choice for a latent diffusion approach is to decide between a learned representation or a manually designed representation. A manually designed representation such as wavelets~\cite{hui2022neural} is easier to design and more lightweight, but in many contexts learned representations have shown to outperform manually designed ones. We therefore opt to explore learned representations. This requires a two-stage training strategy. The first stage is an autoencoder (variational autoencoder) to encode 3D shapes into a latent space. The second stage is training a diffusion model in the learned latent space.

In the case of training diffusion models for 3D neural fields, it is even more necessary to generate in latent space. First, diffusion models often work with data of fixed size (\eg, images of a given fixed resolution). Second, a neural field is a continuous real-valued function that can be seen as an infinite-dimensional vector. For both reasons, we decide to find a way to encode shapes into latent space before all else (as well as a decoding method for reverting latents back to shapes).

\begin{figure*}[tb]
    \centering
    \input{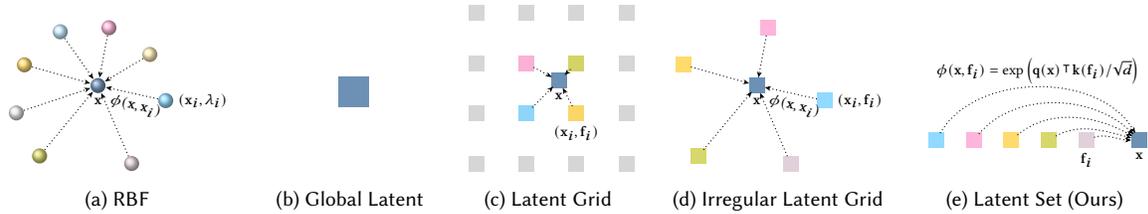}
    \vspace{-10pt}
    \caption{\textbf{Continuous function representations.} Scalars are represented with spheres while vectors are cubes. The arrows show how spatial interpolation is computed. $\mathbf{x}_i$ and $\mathbf{x}$ are the coordinates of an anchor and a querying point respectively. $\lambda_i$ is the SDF value of the anchor point $\mathbf{x}_i$ in (a). $\mathbf{f}_i$ is the associate feature vector located in $\mathbf{x}_i$ in (c)(d). The queried SDF/feature of $\mathbf{x}$ is based on the distance function $\phi(\mathbf{x},\mathbf{x}_i)$ in (a)(c)(d), while our proposed latent set representation (e) utilizes the similarity $\phi(\mathbf{x},\mathbf{f}_i)$ between querying coordinate and anchored features via a cross attention mechanism.}
    \label{fig:con-func}
\end{figure*}

Finally, we have to design a suitable learned neural field representation that provides a good trade-off between compression and reconstruction quality. Such a design typically requires three components: a spatial data structure to store the latent information, a spatial interpolation method, and a neural network architecture. There are multiple options proposed in the literature shown in Fig.~\ref{fig:con-func}.
Early methods used a single global latent vector in combination with an MLP network~\cite{park2019deepsdf, mescheder2019occupancy}. This concept is simple and fast but generally struggles to reconstruct high-quality shapes. 
Better shape details can be achieved by using a 3D regular grid of latents~\cite{peng2020convolutional} together with tri-linear interpolation and an MLP. However, such a representation is too large for generative models and it is only possible to use grids of very low resolution (\eg., $8 \times 8 \times 8)$. By introducing sparsity, \eg.,~\cite{yan2022shapeformer,zhang2022dilg}, latents are arranged in an irregular grid. The latent size is largely reduced, but there is still a lot of room for improvement which we capitalize on in the design of 3DShape2VecSet.

The design of 3DShape2VecSet combines ideas from neural fields, radial basis functions, and the network architecture of attention layers.
Similar to radial basis function representation for continuous functions, we can also re-write existing methods in a similar form (linear combination). Inspired by cross attention in the transformer network~\cite{vaswani2017attention}, we derived the proposed latent representation which is a fixed-size set of latent vectors. There are two main reasons that we believe contribute to the success of the representations. First, the representation is well-suited for the use with transformer-based networks. As transformer-based networks tend to outperform current alternatives, we can better benefit from this network architecture. Instead of only using MLPs to process latent information, we use a linear layer and cross-attention. Second, the representation no longer uses explicitly designed positional features, but only gives the network the option to encode positional information in any form it considers suitable. This is in line with our design principle of favoring learned representations over manually designed ones. See \cref{fig:con-func} e) for the proposed latent representation.

Using our novel shape representation, we can train diffusion models in the learned 3D shape latent space. Our results demonstrate an improved shape encoding quality and generation quality compared to the current state of the art.
While pioneering work in 3D shape generation using diffusion models already showed unconditional 3D shape generation, we show multiple novel applications of 3D diffusion models: category-conditioned generation, text-conditioned shape generation, shape reconstruction from single-view image, and shape reconstruction from partial point clouds.

\smallskip
To sum up, our contributions are as follows:
\begin{enumerate}
    \item We propose a new representation for 3D shapes. Any shape can be represented by a fixed-length array of latents and processed with cross-attention and linear layers to yield a neural field.
    \item We propose a new network architecture to process shapes in the proposed representation, including a building block to aggregate information from a large point cloud using cross-attention.
    \item We improve the state of the art in 3D shape autoencoding to yield a high fidelity reconstruction including local details. 
    \item We propose a latent set diffusion framework that improves the state of the art in 3D shape generation as measured by FID, KID, FPD, and KPD.
    \item We show 3D shape diffusion for category-conditioned generation, text-conditioned generation, point-cloud completion, and image-conditioned generation.
\end{enumerate}

\section{Related Work}

\begin{table}[tb]
\centering
\caption{\textbf{Neural fields for 3D shapes.} We categorize methods according to the position of the latents.}


\begin{tabular}{ccc}
\toprule
\# Latents & Latent Position & Methods \\ \hline
\rowcolor{mytbcol!30} &  & OccNet~\cite{mescheder2019occupancy} \\
\cellcolor{mytbcol!30} & \cellcolor{mytbcol!30} & DeepSDF~\cite{park2019deepsdf} \\
\rowcolor{mytbcol!30}\multirow{-3}{*}{Single} & \multirow{-3}{*}{Global} & IM-Net~\cite{chen2019learning} \\
\hline
\rowcolor{gray!10} &  & ConvOccNet~\cite{peng2020convolutional} \\
\cellcolor{gray!10} & \cellcolor{gray!10}  & IF-Net~\cite{chibane2020implicit} \\
\rowcolor{gray!10} &  & LIG~\cite{jiang2020local}\\ 
\cellcolor{gray!10} & \cellcolor{gray!10} & DeepLS~\cite{chabra2020deep}\\
\rowcolor{gray!10} &  & SA-ConvOccNet~\cite{tang2021sa} \\
\cellcolor{gray!10}\multirow{-6}{*}{Multiple} & \cellcolor{gray!10}\multirow{-6}{*}{Regular Grid} & NKF~\cite{williams2022neural} \\
\hline
\rowcolor{mytbcol!30} & & LDIF~\cite{genova2020local} \\
\cellcolor{mytbcol!30}& \cellcolor{mytbcol!30}& Point2Surf~\cite{erler2020points2surf} \\
\rowcolor{mytbcol!30} &  & DCC-DIF~\cite{li2022learning} \\
\cellcolor{mytbcol!30} & \cellcolor{mytbcol!30} & 3DILG~\cite{zhang2022dilg} \\ 
\rowcolor{mytbcol!30}\multirow{-5}{*}{Multiple}& \multirow{-5}{*}{Irregular Grid} & POCO~\cite{boulch2022poco} \\ \hline
Multiple & Global & Ours \\
\bottomrule
\end{tabular}
\label{table:neural-fields}
\end{table}

\begin{table}[tb]
    \centering
    \caption{\textbf{Generative models for 3d shapes.}}
    \begin{threeparttable}
    \def\arraystretch{1.15}\tabcolsep=0.32em
    \begin{tabular}{ccc}
    \toprule
    & Generative & 3D \\
    & Models & Representation \\
    \hline\hline
    3D-GAN~\cite{wu2016learning} & GAN & Voxels \\
    \rowcolor{mytbcol!30}l-GAN~\cite{achlioptas2018learning} & GAN${}^\star$ & Point Clouds \\
    IM-GAN~\cite{chen2019learning} & GAN${}^\star$ & Fields\\
    \rowcolor{mytbcol!30}PointFlow~\cite{yang2019pointflow} & NF & Point Clouds\\
    GenVoxelNet~\cite{xie2020generative} & EBM & Voxels \\
    \rowcolor{mytbcol!30}PointGrow~\cite{sun2020pointgrow} & AR & Point Clouds \\
    PolyGen~\cite{nash2020polygen} & AR & Meshes \\
    \rowcolor{mytbcol!30}GenPointNet~\cite{xie2021generative} & EBM & Point Clouds \\
    3DShapeGen~\cite{ibing20213d} & GAN${}^\star$ & Fields \\
    \rowcolor{mytbcol!30}DPM~\cite{luo2021diffusion} & DM & Point Clouds \\
    PVD~\cite{zhou20213d} & DM & Point Clouds \\
    \rowcolor{mytbcol!30}AutoSDF\cite{mittal2022autosdf} & AR${}^\star$ & Voxels \\
    CanMap~\cite{cheng2022autoregressive} & AR${}^\star$ & Point Clouds \\
    \rowcolor{mytbcol!30}ShapeFormer\cite{yan2022shapeformer} & AR${}^\star$ & Fields \\
    3DILG~\cite{zhang2022dilg} & AR${}^\star$ & Fields \\
    \rowcolor{mytbcol!30}LION~\cite{zeng2022lion} & DM${}^\star$ & Point Clouds \\ 
    SDF-StyleGAN~\cite{zheng2022sdfstylegan} & GAN & Fields \\
    \rowcolor{mytbcol!30}NeuralWavelet~\cite{hui2022neural} & DM${}^\star$ & Fields \\
    \hline\hline
    TriplaneDiffusion~\cite{shue20223d}${}^\diamond$ & DM${}^\star$ & Fields \\
    \rowcolor{gray!10}DiffusionSDF~\cite{chou2022diffusionsdf}${}^\diamond$ & DM${}^\star$ & Fields \\
    \hline\hline
    Ours & DM${}^\star$ & Fields \\
    \bottomrule
    \end{tabular}
    \begin{tablenotes}
      \small
      \item  ${}^\star$ Generative models in latent space.
      \item ${}^\diamond$ Works in submission.
    \end{tablenotes}
\end{threeparttable}
    \label{table:3d-gen}
\end{table}

In this section, we briefly review the literature of 3D shape learning with various data representations and 3D shape generative models. 

\subsection{3D Shape Representations}
We mainly discuss the following representations for 3D shapes, including voxels, point clouds, and neural fields. 
\paragraph{Voxels.} Voxel grids, extended from 2D pixel grids, simply represent a 3D shape as a discrete volumetric grid. Due to their regular structure, early works take advantage of 3D transposed convolution operators for shape prediction~\cite{wu20153d, choy20163d, wu2016learning,girdhar2016learning, brock2016generative,dai2017shape}. A drawback of the voxels-based decoders is that the computational and memory costs of neural networks cubicly increases with respect to the grid resolution. Thus, most voxel-based methods are limited to low-resolution. Octree-based decoders~\cite{meagher1980octree, tatarchenko2017octree, riegler2017octnet, riegler2017octnetfusion, wang2017cnn, wang2018adaptive, hane2017hierarchical} and sparse hash-based decoders \cite{dai2020sg} take 3D space sparsity into account, alleviating the efficiency issues and supporting high-resolution outputs. 

\paragraph{Point Clouds.} 
Early works on neural-network-based point cloud processing include PointNet~\cite{qi2017pointnet, qi2017pointnet++} and DGCNN~\cite{wang2019dynamic}. These works are built upon per-point fully connected layers. More recently, transformers~\cite{vaswani2017attention} were proposed for point cloud processing, \eg,~\cite{zhao2021point, guo2021pct, zhang2022dilg}. These works are inspired by Vision Transformers (ViT)~\cite{dosovitskiy2020vit} in the image domain. Points are firstly grouped into patches to form tokens and then fed into a transformer with self-attention. In this work, we also introduce a network for processing point clouds. Improving upon previous works, we compress a given point cloud to a small representation that is more suitable for generative modeling.


\paragraph{Neural Fields.} A recent trend is to use neural fields as a 3d data representation. The key building block is a neural network which accepts a 3D coordinate as input, and outputs a scalar~\cite{park2019deepsdf, mescheder2019occupancy,michalkiewicz2019deep,chen2019learning} or a vector~\cite{mildenhall2020nerf, Chan2022}. A 3D object is then implicitly defined by this neural network. Neural fields have gained lots of popularity as they can generate objects with arbitrary topologies and infinite resolution. The methods are also called \emph{neural implicit representations} or \emph{coordinate-based} networks. For neural fields for 3d shape modeling, we can categorize methods into global methods and local methods. 1) The global methods encode a shape with a single global latent vector~\cite{mescheder2019occupancy, park2019deepsdf}. Usually the capacity of these kind of methods is limited and they are unable to encode shape details. 2)
The local methods use localized latent vectors which are defined for 3D positions defined on either a regular~\cite{chibane2020implicit, peng2020convolutional, tang2021sa, jiang2020local} or irregular grid~\cite{genova2020local, li2022learning, zhang2022dilg, boulch2022poco}. In contrast, we propose a latent representation where latent vectors do not have associated 3D positions. Instead, we learn to represent a shape as a list of latent vectors. See \cref{table:neural-fields}.

\subsection{Generative models.}
We have seen great success in different 2D image generative models in the past decade. Popular deep generative methods include generative adversarial networks (GANs)~\cite{goodfellow2014generative}, variational autoencoers (VAEs)~\cite{DBLP:journals/corr/KingmaW13}, normalizing flows (NFs)~\cite{rezende2015variational}, energy-based models~\cite{lecun2006tutorial, xie2016theory}, autoregressive models (ARs)~\cite{van2017neural, esser2021taming} and more recently, diffusion models (DMs)~\cite{ho2020denoising} which are the chosen generative model in our work.



In 3D domain, GANs have been popular for 3D generation~\cite{wu2016learning, achlioptas2018learning, chen2019learning,ibing20213d, zheng2022sdfstylegan}, while only a few works are using NFs~\cite{yang2019pointflow} and VAEs~\cite{mo2019structurenet}. A lot of recent work employs ARs~\cite{sun2020pointgrow, nash2020polygen,mittal2022autosdf,cheng2022autoregressive, yan2022shapeformer,zhang2022dilg}. DMs for 3D shapes are relatively unexplored compared to other generative methods. 

There are several DMs dealing with point cloud data~\cite{luo2021diffusion, zhou20213d, zeng2022lion}. Due to the high freedom degree of regressed coordinates, it is always difficult to obtain clean manifold surfaces via post-processing.
As mentioned before, we believe that neural fields are generally more suitable than point clouds for 3D shape generation.
The area of combining DMs and neural fields is still underexplored. 

DreamFusion~\cite{poole2022dreamfusion} explores how to extract 3D information from a pretrained 2D image diffusion model.
The recent NeuralWavelet~\cite{hui2022neural} first encodes shapes (represented as signed distance fields) into the frequency domain with the wavelet transform, and then train DMs on the frequency coefficients. While this formulation is elegant, generative models generally work better on learned representations. Some concurrent works~\cite{shue20223d, chou2022diffusionsdf} in submission also utilize DMs in a latent space for neural field generation. The TriplaneDiffusion~\cite{shue20223d} trains an autodecoder first for each shape. DiffusionSDF~\cite{chou2022diffusionsdf} runs a shape autoencoder based on triplane features~\cite{peng2020convolutional}.


\paragraph{Summary of 3D generation methods.} We list several 3d generation methods in \cref{table:3d-gen}, highlighting the choice of generative model (GAN, DM, EBM, NF, or AR) and the choice of data structure to represent 3D shapes (point clouds, meshes, voxels or fields). 


\section{Preliminaries}
An attention layer~\cite{vaswani2017attention} has three types of inputs: queries, keys, and values. Queries $\mathbf{Q}=[\mathbf{q}_1, \mathbf{q}_2, \dots, \mathbf{q}_{N_q}]\in\mathbb{R}^{d\times N_{q}}$ and keys $\mathbf{K}=[\mathbf{k}_1, \mathbf{k}_2, \dots, \mathbf{k}_{N_k}]\in\mathbb{R}^{d\times N_{k}}$ are first compared to produce coefficients $\mathbf{q}_j^\intercal\mathbf{k}_i/\sqrt{d}$ (they need to be normalized with the softmax function),
\begin{equation}\label{eq:attn-coef}
    A_{i, j} =
    \frac{\mathbf{q}_j^\intercal\mathbf{k}_i/\sqrt{d}}{\sum^{N_k}_{i=1}\exp\left(
        \mathbf{q}_j^\intercal\mathbf{k}_i/\sqrt{d}.
    \right)}
\end{equation}
The coefficients are then used to (linearly) combine values $\mathbf{V}=[\mathbf{v}_1, \mathbf{v}_2, \dots, \mathbf{v}_{N_k}]\in\mathbb{R}^{d_v\times N_{k}}$. We can write the output of an attention layer as follows,
\begin{equation}\label{eq:attn}
    \begin{aligned}
        &\mathrm{Attention}(\mathbf{Q}, \mathbf{K}, \mathbf{V})\\
        =& 
        \begin{bmatrix}
            \mathbf{o}_{1} & \mathbf{o}_{2} & \cdots & \mathbf{o}_{N_q}
        \end{bmatrix}
        \in\mathbb{R}^{d_v\times N_q}\\
        =&
        \begin{bmatrix}
        \displaystyle\sum^{N_k}_{i=1}  A_{i, 1}\mathbf{v}_i & \displaystyle\sum^{N_k}_{i=1}  A_{i, 2} \mathbf{v}_i & \cdots & \displaystyle\sum^{N_k}_{i=1}  A_{i, N_q} \mathbf{v}_i
        \end{bmatrix}
    \end{aligned}
\end{equation}

\paragraph{Cross Attention.} Given two sets $\mathbf{A} =\left[\mathbf{a}_1, \mathbf{a}_2, \dots, \mathbf{a}_{N_a}\right]\in\mathbb{R}^{d_a\times N_a}$ and $\mathbf{B} =\left[\mathbf{b}_1, \mathbf{b}_2, \dots, \mathbf{b}_{N_b}\right]\in\mathbb{R}^{d_b\times N_b}$, the query vectors $\mathbf{Q}$ are constructed with a linear function $\mathbf{q}(\cdot):\mathbb{R}^{d_a}\rightarrow \mathbb{R}^{d}$ by taking elements of $\mathbf{A}$ as input. Similarly, we construct $\mathbf{K}$ and $\mathbf{V}$ with $\mathbf{k}(\cdot):\mathbb{R}^{d_b}\rightarrow \mathbb{R}^{d}$ and $\mathbf{v}(\cdot):\mathbb{R}^{d_b}\rightarrow \mathbb{R}^{d}$, respectively. The inputs of both $\mathbf{k}(\cdot)$ and $\mathbf{v}(\cdot)$ are from $\mathbf{B}$. Each column in the output of Eq.~\eqref{eq:attn} can be written as,
\begin{equation}\label{eq:cross-attn}
    \mathbf{o}(\mathbf{a}_j, \mathbf{B}) = \sum^{N_b}_{i=1}\mathbf{v}(\mathbf{b}_i) \cdot \frac{1}{Z(\mathbf{a}_j, \mathbf{B})}\exp\left(
        \mathbf{q}(\mathbf{a}_j)^\intercal\mathbf{k}(\mathbf{b}_i)/\sqrt{d}
    \right),
\end{equation}
where $Z(\mathbf{a}_j, \mathbf{B}) = \sum^{N_b}_{i=1}\exp\left(
        \mathbf{q}(\mathbf{a}_j)^\intercal\mathbf{k}(\mathbf{b}_i)/\sqrt{d}
    \right)$ is a normalizing factor.
The cross attention operator between two sets is,
\begin{equation}
    \mathrm{CrossAttn}(\mathbf{A}, \mathbf{B}) = \begin{bmatrix}
    \mathbf{o}(\mathbf{a}_1, \mathbf{B}) & \mathbf{o}(\mathbf{a}_2, \mathbf{B}) &\cdots& \mathbf{o}(\mathbf{a}_{N_a}, \mathbf{B})
    \end{bmatrix}\in\mathbb{R}^{d\times N_a}
\end{equation}

\paragraph{Self Attention.} In the case of self attention, we let the two sets be the same $\mathbf{A} =\mathbf{B}$,
\begin{equation}
    \mathrm{SelfAttn}(\mathbf{A}) = \mathrm{CrossAttn}(\mathbf{A}, \mathbf{A}).
\end{equation}


\begin{figure*}[tb]
    \centering
    \input{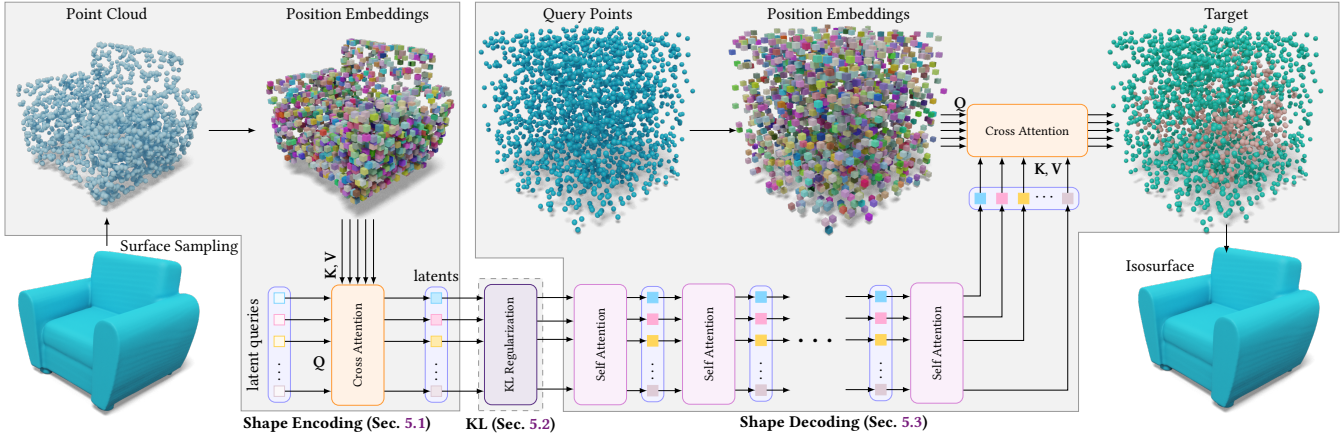}
    \vspace{-20pt}
    \caption{\textbf{Shape autoencoding pipeline.} Given a 3D ground-truth surface mesh as the input, we first sample a point cloud that is mapped to  positional embeddings and encode them into a set of latent codes through a cross-attention module (\textbf{Sec.~\ref{sec:method-shape-enc}}).
    Next, we perform (optional) compression and KL-regularization in the latent space to obtain structured and compact latent shape representations (\textbf{Sec.~\ref{sec:method-kl}}). Finally, the self-attention is carried out to aggregate and exchange the information within the latent set. And a cross-attention module is designed to calculate the interpolation weights of query points. The interpolated feature vectors are fed into a fully connected layer for occupancy prediction (\textbf{Sec.~\ref{sec:method-latent-learning}}).
    }
    \label{fig:auto-pipeline}
\end{figure*}

\section{Latent Representation for Neural Fields}\label{sec:method-latent}

Our representation is inspired by radial basis functions (RBFs). We will therefore describe our surface representation design using RBFs as a starting point, and how we extended them using concepts from neural fields and the transformer architecture.
A continuous function can be represented with a set of weighted points in 3D using RBFs:
\begin{equation}\label{eq:o-rbf}
    \mathcal{\hat{O}}_{\text{RBF}}(\mathbf{x})=\sum^M_{i=1}\textcolor{interp}{\lambda_i}\cdot\textcolor{sim}{\phi(\mathbf{x}, \mathbf{x}_i)}
\end{equation}
where $\phi(\mathbf{x}, \mathbf{x}_i)$ is a radial basis function (RBF) and typically represents the similarity (or dissimilarity) between two inputs,
\begin{equation}
    \phi(\mathbf{x}, \mathbf{x}_i) = \phi(\left\|\mathbf{x}-\mathbf{x}_i\right\|).
\end{equation}
Given ground-truth occupancies of $\mathbf{x}_i$, the values of $\lambda_i$ can be obtained by solving a system of linear equations. In this way, we can represent the continuous function $\mathcal{O}(\cdot)$ as a set of $M$ points including their corresponding weights,
\begin{equation}\left\{\lambda_i\in\mathbb{R}, \mathbf{x}_i\in\mathbb{R}^3\right\}^M_{i=1}.\end{equation}
However, in order to retain the details of a 3d shape, we often need a very large number of points (\eg, $M=80,000$ in~\cite{carr2001reconstruction}). This representation does not benefit from recent advances in representation learning and cannot compete with more compact learned representations. We therefore want to modify the representation to change it into a neural field.

One approach to neural fields is to represent each shape as a separate neural network (making the network weights of a fixed size network the representation of a shape) and train a diffusion process as hypernetwork. A second approach is to have a shared encoder-decoder network for all shapes and represent each shape as a latent computed by the encoder. We opt for the second approach, as it leads to more compact representations because it is jointly learned from all shapes in the data set and the network weights themselves do not count towards the latent representation.
Such a neural field takes a tuple of coordinates $\mathbf{x}$ and $C$-dimensional latent $\mathbf{f}$ as input and outputs occupancy,
\begin{equation}
    \mathcal{\hat{O}}_{\text{NN}}(\mathbf{x}) = \mathrm{NN}(\mathbf{x}, \mathbf{f}),
\end{equation}
where $\mathrm{NN}:\mathbb{R}^3\times \mathbb{R}^C\rightarrow [0, 1]$ is a neural network. A first approach was to use a single global latent $\mathbf{f}$, but a major limitation is the ability to encode shape details~\cite{mescheder2019occupancy}. Some follow-up works study \emph{coordinate-dependent} latents~\cite{peng2020convolutional, chibane2020implicit,sajjadi2022scene} that combine traditional data structures such as regular grids with the neural field concept. Latent vectors are arranged in a spatial data structure and then interpolated (trilinearly) to obtain the coordinate-dependent latent $\mathbf{f}_\mathbf{x}$.
A recent work 3DILG~\cite{zhang2022dilg} proposed a sparse representation for 3D shapes, using latents $\mathbf{f}_i$ arranged in an irregular grid at point locations $\mathbf{x}_i$. The final coordinate-dependent latent $\mathbf{f}_\mathbf{x}$ is then estimated by kernel regression,
\begin{equation}\label{eq:f-3dilg}
    \mathbf{f}_\mathbf{x} = \mathcal{\hat{F}}_{\text{KN}}(\mathbf{x})=\sum^M_{i=1}\textcolor{interp}{\mathbf{f}_i}\cdot
    \textcolor{sim}{\frac{1}{Z\left(
        \mathbf{x}, \left\{\mathbf{x}_i\right\}^M_{i=1}
    \right)}\phi(\mathbf{x}, \mathbf{x}_i)},
\end{equation}
where $Z\left(\mathbf{x}, \left\{\mathbf{x}_i\right\}^M_{i=1}\right) = \sum^M_{i=1}\phi(\mathbf{x}, \mathbf{x}_i)$ is a normalizing factor. Thus the representation for a 3D shape can be written as \begin{equation}
\left\{\mathbf{f}_i\in\mathbb{R}^C, \mathbf{x}_i\in\mathbb{R}^3\right\}^M_{i=1}.
\end{equation}
After that, an $\mathrm{MLP}:\mathbb{R}^C\rightarrow [0, 1]$ is applied to project the approximated feature $\mathcal{\hat{F}}_{\text{KN}}(\mathbf{x})$ to occupancy,
\begin{equation}\label{eq:mlp-3dilg}
\mathcal{\hat{O}}_{\text{3DILG}}(\mathbf{x})=\mathrm{MLP}\left(
    \mathcal{\hat{F}}_{\text{KN}}(\mathbf{x})
\right).
\end{equation}

\paragraph{Neural networks with latent sets (proposed).} We initially explored many variations for 3D shape representation based on irregular and regular grids as well as tri-planes, frequency compositions, and other factored representations. Ultimately, we could not improve on existing irregular grids. However, we were able to achieve a significant improvement with the following change. We aim to keep the structure of an irregular grid and the interpolation, but without representing the actual spatial position explicitly. We let the network encode spatial information.
Both the representations (RBF in Eq.~\eqref{eq:o-rbf} and 3DILG in Eq.~\eqref{eq:f-3dilg}) are composed by two parts, \textcolor{interp}{values} and \textcolor{sim}{similarities}.
We keep the structure of the interpolation, but eliminate explicit point coordinates and integrate cross attention from Eq.~\eqref{eq:cross-attn}. The result is the following \emph{learnable} function approximator,
\begin{equation}
    \label{eq:func-approx-cross-attn}
    \mathcal{\hat{F}}(\mathbf{x})=\sum^M_{i=1}\textcolor{interp}{\mathbf{v}(\mathbf{f}_i)} \cdot \textcolor{sim}{\frac{1}{Z\left(
        \mathbf{x}, \left\{\mathbf{f}_i\right\}^M_{i=1}
    \right)} e^{\mathbf{q}(\mathbf{x})^\intercal\mathbf{k}(\mathbf{f}_i)/\sqrt{d}}},
\end{equation}
where 
$Z\left(\mathbf{x}, \left\{\mathbf{f}_i\right\}^M_{i=1}\right) = \sum^M_{i=1}e^{\mathbf{q}(\mathbf{x})^\intercal\mathbf{k}(\mathbf{f}_i)/\sqrt{d}}$ is a normalizing factor. Similar to the MLP in Eq.~\ref{eq:mlp-3dilg}, we apply a single fully connected layer to get desired occupancy values,
\begin{equation}\label{eq:fc-ours}
\mathcal{\hat{O}}(\mathbf{x}) = \mathrm{FC}\left(\mathcal{\hat{F}}(\mathbf{x})\right).
\end{equation}
Compared to 3DILG and all other coordinate-latent-based methods, we dropped the dependency of the coordinate set $\left\{\mathbf{x}_i\right\}^M_{i=1}$, the new representation only contains a set of latents,
\begin{equation}\label{eq:ours-rep}
\left\{\mathbf{f}_i\in\mathbb{R}^C\right\}^M_{i=1}.
\end{equation}
An alternative view of our proposed function approximator is to see it as cross attention between query points $\mathbf{x}$ and a set of latents.

\section{Network Architecture for Shape Representation Learning}
In this section, we will discuss how we design a variational autoencoder based on the latent representation proposed in \cref{sec:method-latent}. The architecture has three components discussed in the following: a 3D shape encoder, KL regularization block, and a 3D shape decoder.

\subsection{Shape encoding}\label{sec:method-shape-enc}
We sample the surfaces of 3D input shapes in a 3D shape dataset. This results in a point  cloud of size $N$ for each shape, $\{\mathbf{x}_i\in\mathbb{R}^3\}^N_{i=1}$ or in matrix form $\mathbf{X}\in\mathbb{R}^{3\times N}$. 
While the dataset used in the paper originally represents shapes as triangle meshes, our framework is directly compatible with other surface representations, such as scanned point clouds, spline surfaces, or implicit surfaces.

In order to learn representations in the form of \cref{eq:ours-rep}, 
the first challenge is to aggregate the information contained in a possibly large point cloud $\{\mathbf{x}_i\}^N_{i=1}$ into a smaller set of latent vectors $\left\{\mathbf{f}_i\right\}^M_{i=1}$. We design a set-to-set network to this effect. 

\begin{figure}
    \centering
    \subcaptionbox{Learnable Queries}{
        \begin{tikzpicture}
\tiny
\definecolor{olive}{RGB}{255,127,0}

\definecolor{red}{rgb}{0.4,0.8,1}
\definecolor{blue}{rgb}{1,0.6,0.8}
\definecolor{cyan}{RGB}{255,204,51}
\definecolor{yellow}{RGB}{200,200,51}
\definecolor{orange}{RGB}{214,190,204}

\definecolor{aabbee}{rgb}{0.2,0.4,0.6}

\colorlet{mixed1}{red!50!blue}
\colorlet{mixed2}{blue!50!cyan}
\colorlet{mixed3}{cyan!50!yellow}

\tikzset{auto matrix/.style={matrix of nodes,
inner sep=0pt, ampersand replacement=\&,
nodes in empty cells,column sep=1.0pt,row sep=0.4pt,
}}

\node[inner sep=0pt, rectangle, rounded corners, minimum width=1.5cm, minimum height=1.0cm, text centered, draw=olive!80, fill=olive!10,
label={[xshift=-0.8cm, yshift=0.1cm]above:\rotatebox[]{90}{$\mathbf{K}, \mathbf{V}$}},
label={[xshift=-0.1cm, yshift=0.45cm]left:$\mathbf{Q}$}
] (ca) {\tiny Cross Attention};

\matrix[above=0.5cm of ca,auto matrix=n,xshift=0em,yshift=0em,opacity=0.9, 
column 1/.style={nodes={draw=red!80, fill=red!80}},
column 2/.style={nodes={draw=mixed1!80, fill=mixed1!80}},
column 3/.style={nodes={draw=blue!80, fill=blue!80}},
column 4/.style={nodes={draw=mixed2!80, fill=mixed2!80}},
column 5/.style={nodes={draw=cyan!80, fill=cyan!80}},
column 6/.style={nodes={draw=mixed3!80, fill=mixed3!80}},
column 7/.style={nodes={draw=yellow!80, fill=yellow!80}},
column 8/.style={nodes={draw=orange!80, fill=orange!80}},
cells={nodes={minimum width=0.8em,minimum height=0.8em,
very thin,anchor=center,
}}
](n){
\& \& \& \& \& \& \& \\
};

\draw[-stealth] (n-1-1.south) -- (n-1-1.south |- ca.north);
\draw[-stealth] (n-1-2.south) -- (n-1-2.south |- ca.north);
\draw[-stealth] (n-1-3.south) -- (n-1-3.south |- ca.north);
\draw[-stealth] (n-1-4.south) -- (n-1-4.south |- ca.north);
\draw[-stealth] (n-1-5.south) -- (n-1-5.south |- ca.north);
\draw[-stealth] (n-1-6.south) -- (n-1-6.south |- ca.north);
\draw[-stealth] (n-1-7.south) -- (n-1-7.south |- ca.north);
\draw[-stealth] (n-1-8.south) -- (n-1-8.south |- ca.north);

\matrix[left=0.5cm of ca,auto matrix=m,xshift=0em,yshift=0em,opacity=0.9, 
row 1/.style={nodes={draw=red!80, fill=red!10}},
row 2/.style={nodes={draw=blue!80, fill=blue!10}},
row 3/.style={nodes={draw=cyan!80, fill=cyan!10}},
row 4/.style={nodes={draw=yellow!80, fill=yellow!10}},
row 5/.style={nodes={draw=orange!80, fill=orange!10}},
cells={nodes={minimum width=0.8em,minimum height=0.8em,
very thin,anchor=center,
}}
](m){
\\ \\ \\ \\ \\
};

\node[fit=(m), label={left:\rotatebox[]{90}{Learnable}}] () {};

\draw[-stealth] (m-1-1.east) -- (m-1-1.east -| ca.west);
\draw[-stealth] (m-2-1.east) -- (m-2-1.east -| ca.west);
\draw[-stealth] (m-3-1.east) -- (m-3-1.east -| ca.west);
\draw[-stealth] (m-4-1.east) -- (m-4-1.east -| ca.west);
\draw[-stealth] (m-5-1.east) -- (m-5-1.east -| ca.west);

\begin{scope}[]
    \foreach \i in {-2,...,2}{%
      \draw[-stealth] ([yshift=\i * 0.17 cm, xshift=0cm]ca.east) -- ([yshift=\i * 0.17 cm, xshift=0.5cm]ca.east) ;}
\end{scope}

\end{tikzpicture}
    }\quad
    \subcaptionbox{Point Queries}{
        \begin{tikzpicture}
\tiny
\tikzset{myarrow/.style={
    -{Stealth[scale=0.7]},densely dotted,shorten >= 0pt,shorten <= 0pt
    }
}

\definecolor{olive}{RGB}{255,127,0}

\definecolor{red}{rgb}{0.4,0.8,1}
\definecolor{blue}{rgb}{1,0.6,0.8}
\definecolor{cyan}{RGB}{255,204,51}
\definecolor{yellow}{RGB}{200,200,51}
\definecolor{orange}{RGB}{214,190,204}

\definecolor{aabbee}{rgb}{0.2,0.4,0.6}

\colorlet{mixed1}{red!50!blue}
\colorlet{mixed2}{blue!50!cyan}
\colorlet{mixed3}{cyan!50!yellow}

\tikzset{auto matrix/.style={matrix of nodes,
inner sep=0pt, ampersand replacement=\&,
nodes in empty cells,column sep=1.0pt,row sep=0.4pt,
}}

\node[inner sep=0pt, rectangle, rounded corners, minimum width=1.5cm, minimum height=1.0cm, text centered, draw=olive!80, fill=olive!10,
label={[xshift=-0.8cm, yshift=0.1cm]above:\rotatebox[]{90}{$\mathbf{K}, \mathbf{V}$}},
label={[xshift=-0.1cm, yshift=0.45cm]left:$\mathbf{Q}$}
] (ca) {\tiny Cross Attention};

\matrix[above=0.5cm of ca,auto matrix=n,xshift=0em,yshift=0em,opacity=0.9, 
column 1/.style={nodes={draw=red!80, fill=red!80}},
column 2/.style={nodes={draw=mixed1!80, fill=mixed1!80}},
column 3/.style={nodes={draw=blue!80, fill=blue!80}},
column 4/.style={nodes={draw=mixed2!80, fill=mixed2!80}},
column 5/.style={nodes={draw=cyan!80, fill=cyan!80}},
column 6/.style={nodes={draw=mixed3!80, fill=mixed3!80}},
column 7/.style={nodes={draw=yellow!80, fill=yellow!80}},
column 8/.style={nodes={draw=orange!80, fill=orange!80}},
cells={nodes={minimum width=0.8em,minimum height=0.8em,
very thin,anchor=center,
}}
](n){
\& \& \& \& \& \& \& \\
};

\draw[-stealth] (n-1-1.south) -- (n-1-1.south |- ca.north);
\draw[-stealth] (n-1-2.south) -- (n-1-2.south |- ca.north);
\draw[-stealth] (n-1-3.south) -- (n-1-3.south |- ca.north);
\draw[-stealth] (n-1-4.south) -- (n-1-4.south |- ca.north);
\draw[-stealth] (n-1-5.south) -- (n-1-5.south |- ca.north);
\draw[-stealth] (n-1-6.south) -- (n-1-6.south |- ca.north);
\draw[-stealth] (n-1-7.south) -- (n-1-7.south |- ca.north);
\draw[-stealth] (n-1-8.south) -- (n-1-8.south |- ca.north);

\matrix[left=0.5cm of ca,auto matrix=m,xshift=0em,yshift=0em,opacity=0.9, 
row 1/.style={nodes={draw=red!80, fill=red!80}},
row 2/.style={nodes={draw=blue!80, fill=blue!80}},
row 3/.style={nodes={draw=cyan!80, fill=cyan!80}},
row 4/.style={nodes={draw=yellow!80, fill=yellow!80}},
row 5/.style={nodes={draw=orange!80, fill=orange!80}},
cells={nodes={minimum width=0.8em,minimum height=0.8em,
very thin,anchor=center,
}}
](m){
\\ \\ \\ \\ \\
};

\draw[-stealth] (m-1-1.east) -- (m-1-1.east -| ca.west);
\draw[-stealth] (m-2-1.east) -- (m-2-1.east -| ca.west);
\draw[-stealth] (m-3-1.east) -- (m-3-1.east -| ca.west);
\draw[-stealth] (m-4-1.east) -- (m-4-1.east -| ca.west);
\draw[-stealth] (m-5-1.east) -- (m-5-1.east -| ca.west);




\draw[myarrow] (n-1-1.west) -| (m-1-1.north);

\draw[myarrow]  (n-1-3) -- node {} ++(0cm,0.2cm) -- node {} ++(-1.3cm,0cm) |- (m-2-1.west) node[] {} node[] {};

\draw[myarrow]  (n-1-5) -- node {} ++(0cm,0.3cm) -- node {} ++(-1.8cm,0cm) |- (m-3-1.west) node[] {} node[] {};

\draw[myarrow]  (n-1-7) -- node {} ++(0cm,0.4cm) -- node {} ++(-2.3cm,0cm) |- (m-4-1.west) node[] {} node[] {};

\draw[myarrow]  (n-1-8) -- node {} ++(0cm,0.5cm) -- node[midway, above] {Subsample and Copy} ++(-2.6cm,0cm) |- (m-5-1.west) node[] {} node[] {};

\begin{scope}[]
    \foreach \i in {-2,...,2}{%
      \draw[-stealth] ([yshift=\i * 0.17 cm, xshift=0cm]ca.east) -- ([yshift=\i * 0.17 cm, xshift=0.5cm]ca.east) ;}
\end{scope}
\end{tikzpicture}
    }\quad
    \caption{\textbf{Two ways to encode a point cloud.} (a) uses a learnable query set; (b) uses a downsampled version of input point embeddings as the query set.}
    \label{fig:queries}
\end{figure}
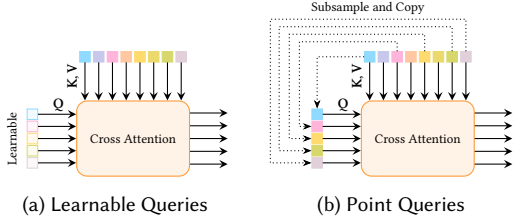
 A popular solution to this problem in previous work is to divide the large point cloud into a smaller set of patches and to learn one latent vector per patch. Although this is a very well researched and standard component in many networks, we discovered a more successful way to aggregate features from a large point cloud that is better compatible with the transformer architecture. We considered two options.

One way is to define a learnable query set. Inspired by DETR~\cite{carion2020end} and Perceiver~\cite{jaegle2021perceiver}, we use the cross attention to encode $\mathbf{X}$,
\begin{equation}\label{eq:enc-learnable}
    \mathrm{Enc}_{\text{learnable}}(\mathbf{X})=\mathrm{CrossAttn}(
        \mathbf{L}, \mathrm{PosEmb}(
            \mathbf{X}
        )
    )\in\mathbb{R}^{C\times M},
\end{equation}
where $\mathbf{L}\in\mathbb{R}^{C\times M}$ is a \emph{learnable query} set where each entry is $C$-dimensional, and $\mathrm{PosEmb}:\mathbb{R}^{3}\rightarrow \mathbb{R}^C$ is a column-wise positional embedding function.

Another way is to utilize the point cloud itself. We first subsample the point cloud $\mathbf{X}$ to a smaller one with furthest point sampling, $\mathbf{X}_0 = \mathrm{FPS}(\mathbf{X})\in\mathbb{R}^{3\times M}$. The cross attention is applied to $\mathbf{X}_0$ and $\mathbf{X}$,
\begin{equation}\label{eq:enc-subsample}
    \mathrm{Enc}_{\text{points}}(\mathbf{X}) = \mathrm{CrossAttn}(\mathrm{PosEmb}(\mathbf{X}_0), \mathrm{PosEmb}(\mathbf{X})),
\end{equation}
which can also be seen as a ``partial'' self attention. See \cref{fig:queries} for an illustration of both design choices. Intuitively, the number $M$ affects the reconstruction performance: the larger the $M$, the better reconstruction. However, $M$ strongly affects the training time due to the transformer architecture, so it should not be too large. In our final model, the number of latents $M$ is set as $512$, and the number of channels $C$ is $512$ to provide a trade off between reconstruction quality and training time.


\subsection{KL regularization block}\label{sec:method-kl}
Latent diffusion~\cite{rombach2022high} proposed to use a variational autoencoder (VAE)~\cite{DBLP:journals/corr/KingmaW13} to compress images. We adapt this design idea for our 3D shape representation and also regularize the latents with KL-divergence.
 We should note that the KL regularization is optional and only necessary for the second-stage diffusion model training. If we just want a method for surface reconstruction from point clouds, we do not need the KL regularization.

We first linear project latents to mean and variance by two network branches, respectively,
\begin{equation}
    \begin{aligned}
        \mathrm{FC}_{\mu}(\mathbf{f}_i) &= \left(\mu_{i, j}\right)_{j\in[1, 2, \cdots, C_0]}\\
        \mathrm{FC}_{\sigma}(\mathbf{f}_i) &= \left(\log\sigma_{i, j}^2\right)_{j\in[1, 2,\cdots,C_0]}
    \end{aligned}
\end{equation}
where $\mathrm{FC}_{\mu}:\mathbb{R}^C\rightarrow\mathbb
R^{C_0}$ and $\mathrm{FC}_{\sigma}:\mathbb{R}^C\rightarrow\mathbb
R^{C_0}$ are two linear projection layers. 
We use a different size of output channels $C_0$, where $C_0 \ll C$. This compression enables us to train diffusion models on smaller latents of total size $M\cdot C_0 \ll M\cdot C$. We can write the bottleneck of the VAE formally, $\forall i\in[1, 2, \cdots, M], j\in[1, 2, \cdots, C_0]$, 
\begin{equation}\label{eq:bottleneck}
    z_{i, j} = \mu_{i,j} + \sigma_{i,j} \cdot \epsilon,
\end{equation}
where $\epsilon\sim\mathcal{N}(0, 1)$.
The KL regularization can be written as,
\begin{equation}
    \begin{aligned}
        \mathcal{L}_{\text{reg}}\left(\{\mathbf{f}_i\}^M_{i=1}\right)=\frac{1}{M\cdot C_0}\sum^{M}_{i=1}\sum^{C_0}_{j=1}\frac{1}{2} \left(\mu_{i,j}^2 + \sigma_{i,j}^2-\log \sigma_{i,j}^2\right).
    \end{aligned}
\end{equation}
In practice, we set the weight for KL loss as $0.001$ and report the performance for different values of $C_0$ in~\cref{sec:result-auto-enc}. Our recommended setting is $C_0 = 32$. 

\subsection{Shape decoding}\label{sec:method-latent-learning}
To increase the expressivity of the network, we add a latent learning network between the two parts. Because our latents are a set of vectors, it is natural to use transformer networks here. Thus, the proposed network here is a series of self attention blocks,

\begin{equation}
    \left\{\mathbf{f}_i\right\}^M_{i=1} \leftarrow \mathrm{SelfAttn}^{(l)}\left(
        \left\{\mathbf{f}_i\right\}^M_{i=1}
    \right), \quad \text{ for } i = 1, \cdots, L.
\end{equation}
The $\mathrm{SelfAttn}(\cdot)$ with a superscript $(l)$ here means $l$-th block. The latents $\{\mathbf{f}_i\}^M_{i=1}$ obtained using either \cref{eq:enc-learnable} or \cref{eq:enc-subsample} are fed into the self attention blocks. 
Given a query $\mathbf{x}$, the corresponding latent is interpolated using \cref{eq:func-approx-cross-attn}, and the occupancy is obtained with a fully connected layer as shown in \cref{eq:fc-ours}.

\paragraph{Loss.} We optimize the binary cross entropy loss between our approximated function and the ground-truth indicator function as in prior works~\cite{mescheder2019occupancy}.
\begin{equation}
\mathcal{L}_{\text{recon}}\left(\{\mathbf{f}_i\}^M_{i=1}, \mathcal{O}\right) = \mathbb{E}_{\mathbf{x}\in\mathbb{R}^3}\left[\mathrm{BCE}\left(
        \mathcal{\hat{O}}(\mathbf{x}), \mathcal{O}(\mathbf{x})
    \right)\right].
\end{equation}

\paragraph{Surface reconstruction.} We sample query points in a grid of resolution $128^3$. The final surface is reconstructed with Marching Cubes~\cite{lorensen1987marching}.

\begin{figure}[tb]
    \centering

\pgfdeclarelayer{background}
\pgfsetlayers{background,main}
    
\tikzset{auto matrix/.style={matrix of nodes,
inner sep=0pt, ampersand replacement=\&,
nodes in empty cells,column sep=1.0pt,row sep=0.4pt,
}}

\definecolor{red}{rgb}{0.4,0.8,1}
\definecolor{blue}{rgb}{1,0.6,0.8}
\definecolor{cyan}{RGB}{255,204,51}
\definecolor{yellow}{RGB}{200,200,51}
\definecolor{orange}{RGB}{214,190,204}

\definecolor{aabbee}{rgb}{0.2,0.6,0.4}

\newcommand{\mymatrix}[6]{
    \let\desc\empty
        \foreach \j in {1,...,#5} {
            \xappto\desc{\&}
        }
        \gappto\desc{\\}

    \matrix[#1,auto matrix=u,xshift=0em,yshift=0em,opacity=0.9, 
    column 1/.style={nodes={draw=red!#2, fill=red!#3}},
    column 2/.style={nodes={draw=blue!#2, fill=blue!#3}},
    column 3/.style={nodes={draw=cyan!#2, fill=cyan!#3}},
    column 4/.style={nodes={draw=yellow!#2, fill=yellow!#3}},
    column 5/.style={nodes={draw=orange!#2, fill=orange!#3}},
    cells={nodes={minimum width=0.8em,minimum height=#4 * 0.8em,
    very thin,anchor=center,
    }}
    ](#6){
    \desc
    };
}

\tiny

\begin{tikzpicture}[tight background, remember picture, baseline={([xshift=0ex]current bounding box.center)}]
    \mymatrix{
        label={[name=f-label]above:$\{\mathbf{f}_i\}^M_{i=1}$}
    }{80}{80}{8}{4}{f}
    \draw[-stealth] ([yshift=-1ex]f.south west) -- 
    ([yshift=-1ex]f.south east) node[midway,below] (f-M) {$M$};
    \draw[-stealth] ([xshift=1ex]f.north east) -- 
    ([xshift=1ex]f.south east) node[midway,right] {$C$};
    \mymatrix{right=1.5cm of f, yshift=0.5cm}{80}{80}{3}{4}{f-mean}
    \draw[-latex] ([xshift=0.15cm, yshift=0.25cm]f.east) -- ([xshift=-0.1cm]f-mean.west) node[xshift=0cm, yshift=0cm, midway, above, sloped] {$\mathrm{FC}_{\mu}$};
    \draw[-stealth] ([yshift=-1ex]f-mean.south west) -- 
    ([yshift=-1ex]f-mean.south east) node[midway,below] {$M$};
    \draw[-stealth] ([xshift=1ex]f-mean.north east) -- 
    ([xshift=1ex]f-mean.south east) node[midway,right] {$C_0$};
    \mymatrix{right=1.5cm of f, yshift=-0.5cm}{80}{80}{3}{4}{f-std}
    \draw[-latex] ([xshift=0.15cm, yshift=-0.25cm]f.east) -- ([xshift=-0.1cm]f-std.west) node[xshift=0cm, yshift=0cm, midway, below, sloped] {$\mathrm{FC}_{\sigma}$};
    \draw[-stealth] ([yshift=-1ex]f-std.south west) -- 
    ([yshift=-1ex]f-std.south east) node[midway,below] (f-std-M){$M$};
    \draw[-stealth] ([xshift=1ex]f-std.north east) -- 
    ([xshift=1ex]f-std.south east) node[midway,right] {$C_0$};
    \mymatrix{right=4cm of f, yshift=0cm, label={above:$\{\mathbf{z}_i\}^M_{i=1}$}}{80}{80}{3}{4}{f-sample}
    \draw[-stealth] ([yshift=-1ex]f-sample.south west) -- 
    ([yshift=-1ex]f-sample.south east) node[midway,below] {$M$};
    \draw[-stealth] ([xshift=1ex]f-sample.north east) -- 
    ([xshift=1ex]f-sample.south east) node[midway,right] {$C_0$};
    \draw[-latex] ([xshift=0.15cm, yshift=-0.25cm]f-mean.east) -- ([xshift=-0.1cm, , yshift=0.1cm]f-sample.west) node[xshift=0cm, yshift=0cm, midway, below, sloped] (mean-arrow) {};
    \draw[-latex] ([xshift=0.15cm, yshift=0.25cm]f-std.east) -- ([xshift=-0.1cm, yshift=-0.1cm]f-sample.west) node[xshift=0cm, yshift=0cm, midway, below, sloped] (std-arrow) {};
    \node [fit=(mean-arrow)(std-arrow)] () {Sample}; 
    \mymatrix{
        label={above:$\{\mathbf{f}_i\}^M_{i=1}$},
        right=6.5cm of f, yshift=0cm
    }{80}{80}{8}{4}{f-recover}
    \draw[-stealth] ([xshift=0.5cm]f-sample.east) -- 
    ([xshift=-0.1cm]f-recover.west) node[midway,below] {$\mathrm{FC}_{\text{up}}$};
    \draw[-stealth] ([yshift=-1ex]f-recover.south west) -- 
    ([yshift=-1ex]f-recover.south east) node[midway,below] {$M$};
    \draw[-stealth] ([xshift=1ex]f-recover.north east) -- 
    ([xshift=1ex]f-recover.south east) node[midway,right] (f-C){$C$};
    \draw[latex-] ([yshift=5ex]f.north) -- node[swap, above, 
        label={[xshift=0.2cm, yshift=0.1cm]above:Shape Encoding (\cref{sec:method-shape-enc})}
      ] (){}  +(90:0.3);
    \draw[-latex] ([yshift=5ex]f-recover.north) -- node[swap, above,
        label={[xshift=-0.2cm, yshift=0.1cm]above:Latent Decoding (\cref{sec:method-latent-learning})}
    ] {}  +(90:0.3);
\end{tikzpicture}
    \vspace{-10pt}
    \caption{\textbf{KL regularization.} Given a set of latents $\{ \mathbf{f}_i \in \mathbb{R}^C \}_{i=1}^M$ obtained from the shape encoding in Sec.~\ref{sec:method-shape-enc}, we employ two linear projection layers $\mathrm{FC}_{\mu}, \mathrm{FC}_{\sigma}$ to predict the mean and variance of a low-dimensional latent space, where a KL regularization commonly used in VAE training is applied to constrain the feature diversity. Then, we obtain smaller latents $\{ \mathbf{z}_i \in \mathbb{R}^{C_0} \}$ of size $M\cdot C_0 \ll M\cdot C$ via reparametrization sampling. Finally, the compressed latents are mapped back to the original space by $\mathrm{FC}_{\text{up}}$ to obtain a higher dimensionality for the shape decoding in Sec.~\ref{sec:method-latent-learning}. }
    \label{fig:gen-pipeline}
\end{figure}
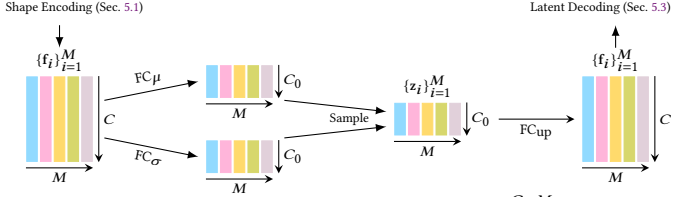

\section{Shape generation}

\begin{figure}
    \centering

\pgfdeclarelayer{background}
\pgfsetlayers{background,main}
    
\tikzset{auto matrix/.style={matrix of nodes,
inner sep=0pt, ampersand replacement=\&,
nodes in empty cells,column sep=1.0pt,row sep=0.4pt,
}}

\definecolor{red}{rgb}{0.4,0.8,1}
\definecolor{blue}{rgb}{1,0.6,0.8}
\definecolor{cyan}{RGB}{255,204,51}
\definecolor{yellow}{RGB}{200,200,51}
\definecolor{orange}{RGB}{214,190,204}

\definecolor{aabbee}{rgb}{0.2,0.6,0.4}

\newcommand{\mymatrix}[6]{
    \let\desc\empty
        \foreach \j in {1,...,#5} {
            \xappto\desc{\&}
        }
        \gappto\desc{\\}

    \matrix[#1,auto matrix=u,xshift=0em,yshift=0em,opacity=0.9, 
    column 1/.style={nodes={draw=red!#2, fill=red!#3}},
    column 2/.style={nodes={draw=blue!#2, fill=blue!#3}},
    column 3/.style={nodes={draw=cyan!#2, fill=cyan!#3}},
    column 4/.style={nodes={draw=yellow!#2, fill=yellow!#3}},
    column 5/.style={nodes={draw=orange!#2, fill=orange!#3}},
    cells={nodes={minimum width=0.8em,minimum height=#4 * 0.8em,
    very thin,anchor=center,
    }}
    ](#6){
    \desc
    };
}

\tiny

\begin{tikzpicture}[tight background, remember picture, baseline={([xshift=0ex]current bounding box.center)}]

    \mymatrix{xshift=0cm, yshift=-2cm}{80}{80}{3}{4}{f-enc-0}
    \mymatrix{right=1.5cm of f-enc-0}{80}{50}{3}{4}{f-enc-1}
    \mymatrix{right=4cm of f-enc-0}{80}{25}{3}{4}{f-enc-2}
    \mymatrix{right=6.5cm of f-enc-0}{80}{10}{3}{4}{f-enc-3}
    \draw[-latex] ([xshift=0.1cm]f-enc-0.east) -- 
    ([xshift=-0.1cm]f-enc-1.west) node[midway,above] {Add Noise};
    \draw[-latex] ([xshift=0.1cm]f-enc-1.east) -- 
    ([xshift=-0.1cm]f-enc-2.west) node[midway,above] {Add Noise};
    \draw[-latex] ([xshift=0.1cm]f-enc-2.east) -- 
    ([xshift=-0.1cm]f-enc-3.west) node[midway,above] {Add Noise};
    \begin{pgfonlayer}{background}
    \node [draw, draw=gray!80, fill=gray!10, fit=(f-enc-0)(f-enc-3), label={[xshift=0cm]above:Forward Diffusion Process}] (destroy) {}; 
    \end{pgfonlayer}
    \mymatrix{below=0.7cm of f-enc-0}{80}{80}{3}{4}{f-dec-0}
    \mymatrix{right=1.5cm of f-dec-0}{80}{50}{3}{4}{f-dec-1}
    \mymatrix{right=4cm of f-dec-0}{80}{25}{3}{4}{f-dec-2}
    \mymatrix{right=6.5cm of f-dec-0}{80}{10}{3}{4}{f-dec-3}
    \draw[latex-] ([xshift=0.1cm]f-dec-0.east) -- 
    ([xshift=-0.1cm]f-dec-1.west) node[midway,above] {Denoise};
    \draw[latex-] ([xshift=0.1cm]f-dec-1.east) -- 
    ([xshift=-0.1cm]f-dec-2.west) node[midway,above] {Denoise};
    \draw[latex-] ([xshift=0.1cm]f-dec-2.east) -- 
    ([xshift=-0.1cm]f-dec-3.west) node[midway,above] {Denoise};
    \begin{pgfonlayer}{background}
    \node [draw, draw=gray!80, fill=gray!10, fit=(f-dec-0)(f-dec-3), label={[xshift=0cm]above:Reverse Diffusion Process}] (recover) {}; 
    \end{pgfonlayer}
    \node[draw, draw=aabbee!80, fill=aabbee!40, right=0.6cm of $(f-enc-3)!0.5!(f-dec-3)$] (cond) {\rotatebox{90}{Condition}};
    \draw[-latex, densely dotted]  (cond) -- node {} ++(0,-1.1cm) -| (f-dec-3) node[] {} node[] {};
    \draw[-latex, densely dotted]  (cond) -- node {} ++(0,-1.1cm) -| (f-dec-2) node[] {} node[] {};
    \draw[-latex, densely dotted]  (cond) -- node {} ++(0,-1.1cm) -| (f-dec-1) node[] {} node[] {};
    \draw[-latex, densely dotted]  (cond) -- node {} ++(0,-1.1cm) -| (f-dec-0) node[] {} node[] {};
\end{tikzpicture}
    \vspace{-10pt}
    \caption{\textbf{Latent set diffusion models.} The diffusion model operates on compressed 3D shapes in the form of a regularized set of latent vectors $\{ \mathbf{z}_i \}_{i=1}^{M}$.}
    \label{fig:diffusion-forward-reverse}
\end{figure}
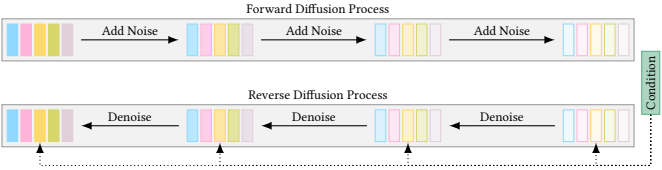

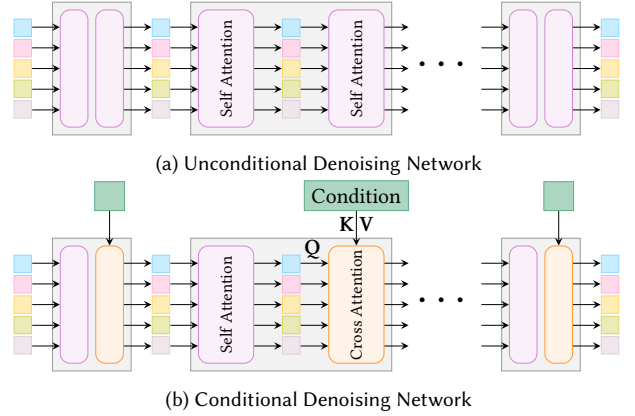
\begin{figure}[tb]
    \centering
    \subcaptionbox{Unconditional Denoising Network}{
        \resizebox{0.95\linewidth}{!}{%
\begin{tikzpicture}[tight background, remember picture]
    \pgfdeclarelayer{background}
    \pgfsetlayers{background,main}
    
    \definecolor{wwccqq}{rgb}{0.4,0.8,0}
    \definecolor{wwccff}{rgb}{0.4,0.8,1}
    \definecolor{ffzzcc}{rgb}{1,0.6,0.8}
    \definecolor{orange}{RGB}{255,204,51}
    \definecolor{violet}{RGB}{125,125,255}
    \definecolor{olive}{RGB}{255,127,0}
    \definecolor{mycolor}{RGB}{214,190,204}
    
    \definecolor{aabbcc}{rgb}{0.8,0.4,0.8}
    \definecolor{aabbdd}{rgb}{0.2,0.2,0}
    \definecolor{aabbee}{rgb}{0.2,0.6,0.8}

    \tikzstyle{attention} = [inner sep=0pt, rectangle, rounded corners, minimum width=0.8cm, minimum height=1.7cm, text centered, draw=olive!80, fill=olive!10]

\tikzset{auto matrix/.style={matrix of nodes,
inner sep=0pt, ampersand replacement=\&,
nodes in empty cells,column sep=0.4pt,row sep=1.0pt,
}}

\definecolor{red}{rgb}{0.4,0.8,1}
\definecolor{blue}{rgb}{1,0.6,0.8}
\definecolor{cyan}{RGB}{255,204,51}
\definecolor{yellow}{RGB}{200,200,51}
\definecolor{orange}{RGB}{214,190,204}

\definecolor{aabbee}{rgb}{0.2,0.6,0.4}

\newcommand{\mymatrix}[6]{
    \let\desc\empty
        \foreach \j in {1,...,#5} {
            \xappto\desc{\\}
        }

    \matrix[#1,auto matrix=u,xshift=0em,yshift=0em,opacity=0.9, 
    row 1/.style={nodes={draw=red!#2, fill=red!#3}},
    row 2/.style={nodes={draw=blue!#2, fill=blue!#3}},
    row 3/.style={nodes={draw=cyan!#2, fill=cyan!#3}},
    row 4/.style={nodes={draw=yellow!#2, fill=yellow!#3}},
    row 5/.style={nodes={draw=orange!#2, fill=orange!#3}},
    cells={nodes={minimum width=#4 * 0.8em,minimum height=0.8em,
    very thin,anchor=center,
    }}
    ](#6){
    \desc
    };
}

    \mymatrix{}{80}{40}{1}{5}{i}

    
    \node[right=0.4cm of i, attention, draw=aabbcc!80, fill=aabbcc!10, minimum width=0.4cm] (self-attn1) {};

    \node[right=0.1cm of self-attn1, attention, draw=aabbcc!80, fill=aabbcc!10, minimum width=0.4cm] (self-attn2) {};

    \begin{pgfonlayer}{background}
    \node [draw, draw=gray!80, fill=gray!10, fit=(self-attn1)(self-attn2)] {}; 
    \end{pgfonlayer}
    
    \begin{scope}[]
        \foreach \i in {-2,...,2}{%
          \draw[-stealth] ([yshift=\i * 0.3 cm, xshift=0cm]i.east) -- ([yshift=\i * 0.3 cm, xshift=0cm]self-attn1.west) ;}
    \end{scope}

    \mymatrix{right=0.4cm of self-attn2,}{80}{40}{1}{5}{f1}

    \begin{scope}[]
        \foreach \i in {-2,...,2}{%
          \draw[-stealth] ([yshift=\i * 0.3 cm, xshift=0cm]self-attn2.east) -- ([yshift=\i * 0.3 cm, xshift=0cm]f1.west) ;}
    \end{scope}

    \node[right=0.4cm of f1, attention, draw=aabbcc!80, fill=aabbcc!10] (self-attn3) {\rotatebox{90}{\footnotesize Self Attention}};

    \begin{scope}[]
        \foreach \i in {-2,...,2}{%
          \draw[-stealth] ([yshift=\i * 0.3 cm, xshift=0cm]f1.east) -- ([yshift=\i * 0.3 cm, xshift=0cm]self-attn3.west) ;}
    \end{scope}
    
    \mymatrix{right=0.4cm of self-attn3,}{80}{40}{1}{5}{f2}

    \begin{scope}[]
        \foreach \i in {-2,...,2}{%
          \draw[-stealth] ([yshift=\i * 0.3 cm, xshift=0cm]self-attn3.east) -- ([yshift=\i * 0.3 cm, xshift=0cm]f2.west) ;}
    \end{scope}

    \node[right=0.4cm of f2, attention, draw=aabbcc!80, fill=aabbcc!10] (self-attn4) {\rotatebox{90}{\footnotesize Self Attention}};
    
    \begin{scope}[]
        \foreach \i in {-2,...,2}{%
          \draw[-stealth] ([yshift=\i * 0.3 cm, xshift=0cm]f2.east) -- ([yshift=\i * 0.3 cm, xshift=0cm]self-attn4.west) ;}
    \end{scope}

    \node[right=0.35cm of self-attn4] (cdots) {\scalebox{2}{$\cdots$}};

    \begin{scope}[]
        \foreach \i in {-2,...,2}{%
          \draw[-stealth] ([yshift=\i * 0.3 cm, xshift=0cm]self-attn4.east) -- ([yshift=\i * 0.3 cm, xshift=0cm]cdots.west) ;}
    \end{scope}

    \begin{pgfonlayer}{background}
    \node [draw, draw=gray!80, fill=gray!10, fit=(self-attn3)(self-attn4)] {}; 
    \end{pgfonlayer}
    
    \node[right=0.4cm of cdots, attention, draw=aabbcc!80, fill=aabbcc!10, minimum width=0.4cm] (self-attn5) {};

    \node[right=0.1cm of self-attn5, attention, draw=aabbcc!80, fill=aabbcc!10, minimum width=0.4cm] (self-attn6) {};

    \begin{pgfonlayer}{background}
    \node [draw, draw=gray!80, fill=gray!10, fit=(self-attn5)(self-attn6)] {}; 
    \end{pgfonlayer}
    
    \begin{scope}[]
        \foreach \i in {-2,...,2}{%
          \draw[-stealth] ([yshift=\i * 0.3 cm, xshift=0cm]cdots.east) -- ([yshift=\i * 0.3 cm, xshift=0cm]self-attn5.west) ;}
    \end{scope}

    \mymatrix{right=0.4cm of self-attn6,}{80}{40}{1}{5}{f3}

    \begin{scope}[]
        \foreach \i in {-2,...,2}{%
          \draw[-stealth] ([yshift=\i * 0.3 cm, xshift=0cm]self-attn6.east) -- ([yshift=\i * 0.3 cm, xshift=0cm]f3.west) ;}
    \end{scope}
\end{tikzpicture}
}
    }\quad
    \subcaptionbox{Conditional Denoising Network}{
        \resizebox{0.95\linewidth}{!}{%
\begin{tikzpicture}[tight background, remember picture]
    \pgfdeclarelayer{background}
    \pgfsetlayers{background,main}
    
    \definecolor{wwccqq}{rgb}{0.4,0.8,0}
    \definecolor{wwccff}{rgb}{0.4,0.8,1}
    \definecolor{ffzzcc}{rgb}{1,0.6,0.8}
    \definecolor{orange}{RGB}{255,204,51}
    \definecolor{violet}{RGB}{125,125,255}
    \definecolor{olive}{RGB}{255,127,0}
    \definecolor{mycolor}{RGB}{214,190,204}
    
    \definecolor{aabbcc}{rgb}{0.8,0.4,0.8}
    \definecolor{aabbdd}{rgb}{0.2,0.2,0}
    \definecolor{aabbee}{rgb}{0.2,0.6,0.8}

    \tikzstyle{attention} = [inner sep=0pt, rectangle, rounded corners, minimum width=0.8cm, minimum height=1.7cm, text centered, draw=olive!80, fill=olive!10]

\tikzset{auto matrix/.style={matrix of nodes,
inner sep=0pt, ampersand replacement=\&,
nodes in empty cells,column sep=0.4pt,row sep=1.0pt,
}}

\definecolor{red}{rgb}{0.4,0.8,1}
\definecolor{blue}{rgb}{1,0.6,0.8}
\definecolor{cyan}{RGB}{255,204,51}
\definecolor{yellow}{RGB}{200,200,51}
\definecolor{orange}{RGB}{214,190,204}

\definecolor{aabbee}{rgb}{0.2,0.6,0.4}

\newcommand{\mymatrix}[6]{
    \let\desc\empty
        \foreach \j in {1,...,#5} {
            \xappto\desc{\\}
        }

    \matrix[#1,auto matrix=u,xshift=0em,yshift=0em,opacity=0.9, 
    row 1/.style={nodes={draw=red!#2, fill=red!#3}},
    row 2/.style={nodes={draw=blue!#2, fill=blue!#3}},
    row 3/.style={nodes={draw=cyan!#2, fill=cyan!#3}},
    row 4/.style={nodes={draw=yellow!#2, fill=yellow!#3}},
    row 5/.style={nodes={draw=orange!#2, fill=orange!#3}},
    cells={nodes={minimum width=#4 * 0.8em,minimum height=0.8em,
    very thin,anchor=center,
    }}
    ](#6){
    \desc
    };
}

    \mymatrix{}{80}{40}{1}{5}{i}

    
    \node[right=0.4cm of i, attention, draw=aabbcc!80, fill=aabbcc!10, minimum width=0.4cm] (self-attn1) {};

    \node[right=0.1cm of self-attn1, attention, minimum width=0.4cm] (self-attn2) {};

    \begin{pgfonlayer}{background}
    \node [draw, draw=gray!80, fill=gray!10, fit=(self-attn1)(self-attn2)] {}; 
    \end{pgfonlayer}
    
    \begin{scope}[]
        \foreach \i in {-2,...,2}{%
          \draw[-stealth] ([yshift=\i * 0.3 cm, xshift=0cm]i.east) -- ([yshift=\i * 0.3 cm, xshift=0cm]self-attn1.west) ;}
    \end{scope}

    \mymatrix{right=0.4cm of self-attn2,}{80}{40}{1}{5}{f1}

    \begin{scope}[]
        \foreach \i in {-2,...,2}{%
          \draw[-stealth] ([yshift=\i * 0.3 cm, xshift=0cm]self-attn2.east) -- ([yshift=\i * 0.3 cm, xshift=0cm]f1.west) ;}
    \end{scope}

    \node[right=0.4cm of f1, attention, draw=aabbcc!80, fill=aabbcc!10] (self-attn3) {\rotatebox{90}{\footnotesize Self Attention}};

    \begin{scope}[]
        \foreach \i in {-2,...,2}{%
          \draw[-stealth] ([yshift=\i * 0.3 cm, xshift=0cm]f1.east) -- ([yshift=\i * 0.3 cm, xshift=0cm]self-attn3.west) ;}
    \end{scope}
    
    \mymatrix{right=0.4cm of self-attn3,}{80}{40}{1}{5}{f2}

    \begin{scope}[]
        \foreach \i in {-2,...,2}{%
          \draw[-stealth] ([yshift=\i * 0.3 cm, xshift=0cm]self-attn3.east) -- ([yshift=\i * 0.3 cm, xshift=0cm]f2.west) ;}
    \end{scope}

    \node[right=0.4cm of f2, attention,
    label={[xshift=0cm, yshift=0.1cm]above:$\mathbf{K}\ \mathbf{V}$},
    label={[xshift=0cm, yshift=0.8cm]left:$\mathbf{Q}$}
    ] (self-attn4) {\rotatebox{90}{\footnotesize Cross Attention}};
    
    \begin{scope}[]
        \foreach \i in {-2,...,2}{%
          \draw[-stealth] ([yshift=\i * 0.3 cm, xshift=0cm]f2.east) -- ([yshift=\i * 0.3 cm, xshift=0cm]self-attn4.west) ;}
    \end{scope}

    \node[right=0.35cm of self-attn4] (cdots) {\scalebox{2}{$\cdots$}};

    \begin{scope}[]
        \foreach \i in {-2,...,2}{%
          \draw[-stealth] ([yshift=\i * 0.3 cm, xshift=0cm]self-attn4.east) -- ([yshift=\i * 0.3 cm, xshift=0cm]cdots.west) ;}
    \end{scope}

    \begin{pgfonlayer}{background}
    \node [draw, draw=gray!80, fill=gray!10, fit=(self-attn3)(self-attn4)] {}; 
    \end{pgfonlayer}

    \node[right=0.4cm of cdots, attention, draw=aabbcc!80, fill=aabbcc!10, minimum width=0.4cm] (self-attn5) {};

    \node[right=0.1cm of self-attn5, attention,minimum width=0.4cm] (self-attn6) {};

    \begin{pgfonlayer}{background}
    \node [draw, draw=gray!80, fill=gray!10, fit=(self-attn5)(self-attn6)] {}; 
    \end{pgfonlayer}

    \begin{scope}[]
        \foreach \i in {-2,...,2}{%
          \draw[-stealth] ([yshift=\i * 0.3 cm, xshift=0cm]cdots.east) -- ([yshift=\i * 0.3 cm, xshift=0cm]self-attn5.west) ;}
    \end{scope}

    \mymatrix{right=0.4cm of self-attn6,}{80}{40}{1}{5}{f3}

    \begin{scope}[]
        \foreach \i in {-2,...,2}{%
          \draw[-stealth] ([yshift=\i * 0.3 cm, xshift=0cm]self-attn6.east) -- ([yshift=\i * 0.3 cm, xshift=0cm]f3.west) ;}
    \end{scope}

    \node[draw, draw=aabbee!80, fill=aabbee!40, above=0.5cm of self-attn2] (cond1) {\phantom A};

    \draw[-stealth] ([]cond1.south) -- ([]self-attn2.north) ;

    \node[draw, draw=aabbee!80, fill=aabbee!40, above=0.5cm of self-attn4] (cond2) {Condition};

    \draw[-stealth] ([]cond2.south) -- ([]self-attn4.north) ;

    \node[draw, draw=aabbee!80, fill=aabbee!40, above=0.5cm of self-attn6] (cond3) {\phantom A};

    \draw[-stealth] ([]cond3.south) -- ([]self-attn6.north) ;
    
\end{tikzpicture}
}
    }\quad
    \vspace{-10pt}
    \caption{
    \textbf{Denoising network}. Our denoising network is composed of several denoising layers (a box in the figure denotes a layer). The denoising layer for unconditional generation contains two sequential self attention blocks. The denoising layer for conditional generation contains a self attention and a cross attention block. The cross attention is for injecting condition information such as categories, images or partial point clouds.
    }
    \label{fig:denoising-layer}
\end{figure}

Our proposed diffusion model combines design decisions from latent diffusion (the idea of the compressed latent space), EDM~\cite{Karras2022edm} (most of the training details), and our shape representation design (the architecture is based on attention and self-attention instead of convolution).

We train diffusion models in the latent space, \ie, the bottleneck in \cref{eq:bottleneck}.
Following the diffusion formulation in EDM~\cite{Karras2022edm}, our denoising objective is
\begin{equation}
    \mathbb{E}_{\mathbf{n}_i\sim\mathcal{N}(\mathbf{0}, \sigma^2\mathbf{I})}
    \frac{1}{M}\sum^M_{i=1}
    \left\|\mathrm{Denoiser}\left(
        \{\mathbf{z}_i+\mathbf{n}_i\}^M_{i=1}, \sigma, \mathcal{C}
    \right)_i-\mathbf{z}_i
    \right\|^2_2,
\end{equation}
where $\mathrm{Denoiser}(\cdot, \cdot, \cdot)$ is our denoising neural network, $\sigma$ is the noise level, and $\mathcal{C}$ is the optional conditional information (\eg, categories, images, partial point clouds and texts). We denote the corresponding output of $\mathbf{z}_i + \mathbf{n}_i$ with the subscript $i$, i.e. $\mathrm{Denoiser}(\cdot, \cdot, \cdot)_i$. We should minimize the loss for every noise level $\sigma$. The sampling is done by solving ordinary/stochastic differential equations (ODE/SDE). 
See \cref{fig:diffusion-forward-reverse} for an illustration and EDM~\cite{Karras2022edm} for a detailed description for both the forward (training) and reverse (sampling) process.

The function $\mathrm{Denoiser}(\cdot, \cdot, \cdot)$ is a set denoising network (set-to-set function). The network can be easily modeled by a self-attention transformer. 
Each layer consists of two attention blocks. The first one is a self attention for attentive learning of the latent set. The second one is for injecting the condition information $\mathcal{C}$ (\cref{fig:denoising-layer} (b)) as in prior works~\cite{rombach2022high}. For simple information like categories, $\mathcal{C}$ is a learnable embedding vector (\eg, 55 different embedding vectors for 55 categories). 
For a single-view image , we use ResNet-18~\cite{he2016deep} as the context encoder to extract a global feature vector as condition $\mathcal{C}$.
For text conditioning, we use BERT~\cite{devlin2018bert} to learn a global feature vector as $\mathcal{C}$. For partial point clouds, we use the shape encoder introduced in Sec.~\ref{sec:method-shape-enc} to obtain a set of latent embeddings as $\mathcal{C}$.
In the case of unconditional generation, the cross attention degrades to self attention (\cref{fig:denoising-layer} (a)). 




\section{Experimental Setup}

\begin{table*}[]
\centering
\caption{\textbf{Shape autoencoding (surface reconstruction from point clouds) on ShapeNet.} We show averaged metrics on all 55 categories and individual metrics for the 7 largest categories. We compare with existing representative methods, \textbf{OccNet} (global latent), \textbf{ConvOccNet} (local latent grid), \textbf{IF-Net} (multiscale local latent grid), and \textbf{3DILG} (irregular latent grid).
For our method, we show two different designs. The column \textbf{Learned Queries} shows results of using~\cref{eq:enc-learnable}, while the column \textbf{Point Queries} means we are using a subsampled point set as queries in~\cref{eq:enc-subsample}. The results of \textbf{Point Queries} are generally better than \textbf{Learned Queries}. This is expected because input-dependent queries (\textbf{Point Queries}) are better than fixed queries (\textbf{Learned Queries}).}
\vspace{-10pt}
\begin{tabular}{cccccccc}
\toprule
 &          & \multirow{2}{*}{OccNet}  & \multirow{2}{*}{ConvOccNet} & \multirow{2}{*}{IF-Net}          & \multirow{2}{*}{3DILG}            & \multicolumn{2}{c}{Ours}            \\ 
 &          &       &       &       &                &      Learned Queries      &      Point Queries      \\ \hline
\rowcolor{mytbcol!30}
 & table    & 0.823 & 0.847 & 0.901 & 0.963          & 0.965          & \textbf{0.971} \\ 
\cellcolor{mytbcol!30} & \cellcolor{mytbcol!30}car      & 0.911 & 0.921 & 0.952 & 0.961          & 0.966          & \textbf{0.969} \\
\rowcolor{mytbcol!30} & chair    & 0.803 & 0.856 & 0.927 & 0.950          & 0.957          & \textbf{0.964} \\
\cellcolor{mytbcol!30} & \cellcolor{mytbcol!30}airplane & 0.835 & 0.881 & 0.937 & 0.952          & 0.962          & \textbf{0.969} \\
\rowcolor{mytbcol!30} & sofa     & 0.894 & 0.930 & 0.960 & 0.975          & 0.975          & \textbf{0.982} \\
\cellcolor{mytbcol!30} & \cellcolor{mytbcol!30}rifle    & 0.755 & 0.871 & 0.914 & 0.938          & 0.947          & \textbf{0.960} \\
\rowcolor{mytbcol!30} & lamp     & 0.735 & 0.859 & 0.914 & 0.926          & 0.931          & \textbf{0.956} \\ 
\cline{3-8}
\cellcolor{mytbcol!30} &  \cellcolor{mytbcol!30}mean (selected) & 0.822 & 0.881 & 0.929 & 0.952          & 0.957          & \textbf{0.967} \\
\rowcolor{mytbcol!30} \multirow{-9}{*}{\cellcolor{mytbcol!30} IoU $\uparrow$} & mean (all) & 0.825 & 0.888 & 0.934 & 0.953          & 0.955          & \textbf{0.965} \\ \hline
\rowcolor{gray!10}  & table    & 0.041 & 0.036 & 0.029 & \textbf{0.026} & \textbf{0.026} & \textbf{0.026} \\
\cellcolor{gray!10} & \cellcolor{gray!10}car      & 0.082 & 0.083 & 0.067 & 0.066          & \textbf{0.062} & \textbf{0.062} \\
\rowcolor{gray!10}  & chair    & 0.058 & 0.044 & 0.031 & 0.029          & 0.028          & \textbf{0.027} \\
\cellcolor{gray!10} & \cellcolor{gray!10}airplane & 0.037 & 0.028 & 0.020 & 0.019          & 0.018          & \textbf{0.017} \\
\rowcolor{gray!10} & sofa     & 0.051 & 0.042 & 0.032 & 0.030          & 0.030          & \textbf{0.029} \\
\cellcolor{gray!10} & \cellcolor{gray!10}rifle    & 0.046 & 0.025 & 0.018 & 0.017          & 0.016          & \textbf{0.014} \\
\rowcolor{gray!10}  & lamp     & 0.090 & 0.050 & 0.038 & 0.036          & 0.035          & \textbf{0.032} \\ 
 \cline{3-8}
\cellcolor{gray!10} & \cellcolor{gray!10}mean (selected)  & 0.058 & 0.040 & 0.034 & 0.032          & 0.031          & \textbf{0.030} \\
\rowcolor{gray!10} \multirow{-9}{*}{Chamfer $\downarrow$} & mean (all)   & 0.072 & 0.052 & 0.041 & 0.040          & 0.039          & \textbf{0.038} \\ \hline
\rowcolor{mytbcol!30} & table    & 0.961 & 0.982 & 0.998 & \textbf{0.999} & \textbf{0.999} & \textbf{0.999} \\
\cellcolor{mytbcol!30} & \cellcolor{mytbcol!30}car      & 0.830 & 0.852 & 0.888 & 0.892          & 0.898          & \textbf{0.899} \\
\rowcolor{mytbcol!30} & chair    & 0.890 & 0.943 & 0.990 & 0.992          & 0.994          & \textbf{0.997} \\
\cellcolor{mytbcol!30} & \cellcolor{mytbcol!30}airplane & 0.948 & 0.982 & 0.994 & 0.993          & 0.994          & \textbf{0.995} \\
\rowcolor{mytbcol!30} & sofa     & 0.918 & 0.967 & 0.988 & 0.986          & 0.986          & \textbf{0.990} \\
\cellcolor{mytbcol!30} & \cellcolor{mytbcol!30}rifle    & 0.922 & 0.987 & 0.998 & 0.997          & 0.998          & \textbf{0.999} \\
\rowcolor{mytbcol!30} & lamp     & 0.820 & 0.945 & 0.970 & 0.971          & 0.970          & \textbf{0.975} \\ 
 \cline{3-8}
\cellcolor{mytbcol!30} &  \cellcolor{mytbcol!30}mean (selected)  & 0.898 & 0.951 & 0.975 & 0.976          & 0.977          & \textbf{0.979} \\
\rowcolor{mytbcol!30}\multirow{-9}{*}{\cellcolor{mytbcol!30} F-Score $\uparrow$} &  mean (all)  & 0.858 & 0.933 & 0.967 & 0.966          & 0.966          & \textbf{0.970} \\ 
\bottomrule
\end{tabular}
\label{table:main}
\end{table*}

\begin{figure*}[h]
    \vspace*{10pt}
    \centering
    \begin{overpic}[trim={0cm 0cm 0cm 0cm},clip,width=1\linewidth,grid=false]{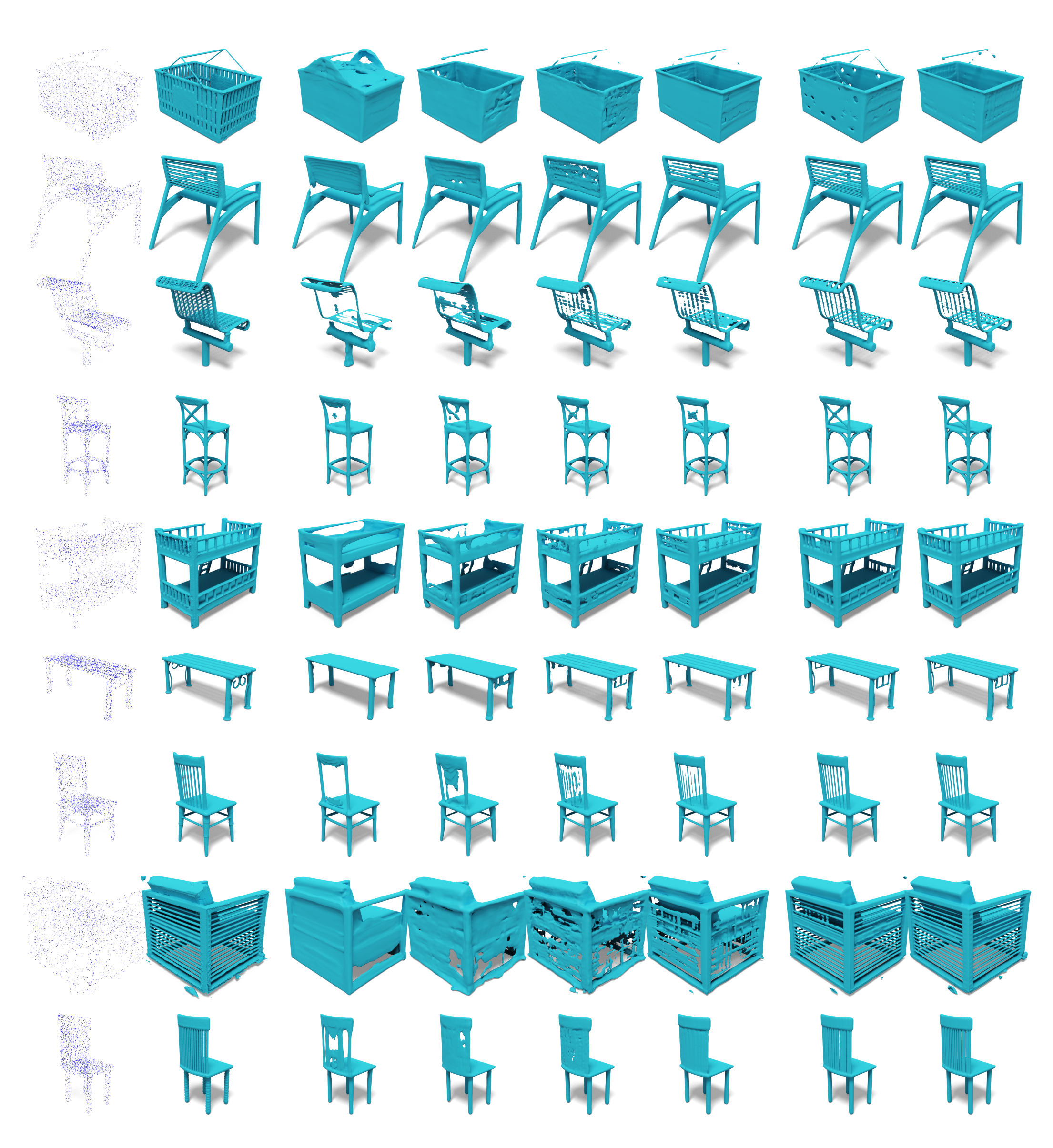}
        \put(5, 98){\small{Input}}
        \put(17, 98){\small{GT}}
        \dashline{0.7}(24.5,3)(24.5,98)
        \dashline{0.7}(68,3)(68,98)
        \put(28, 98){\small{OccNet}}
        \put(38, 98){\small{ConvONet}}
        \put(49, 98){\small{IF-Net}}
        \put(59, 98){\small{3DILG}}
        \put(77, 99){\small{Proposed}}
        \put(69, 97){\small{Learnable Queries}}
        \put(81, 97){\small{Point Queries}}
    \end{overpic}
    \caption{\textbf{
    Visualization of shape autoencoding results (surface reconstruction from point clouds from ShapeNet).}}
    \label{fig:main}
\end{figure*}



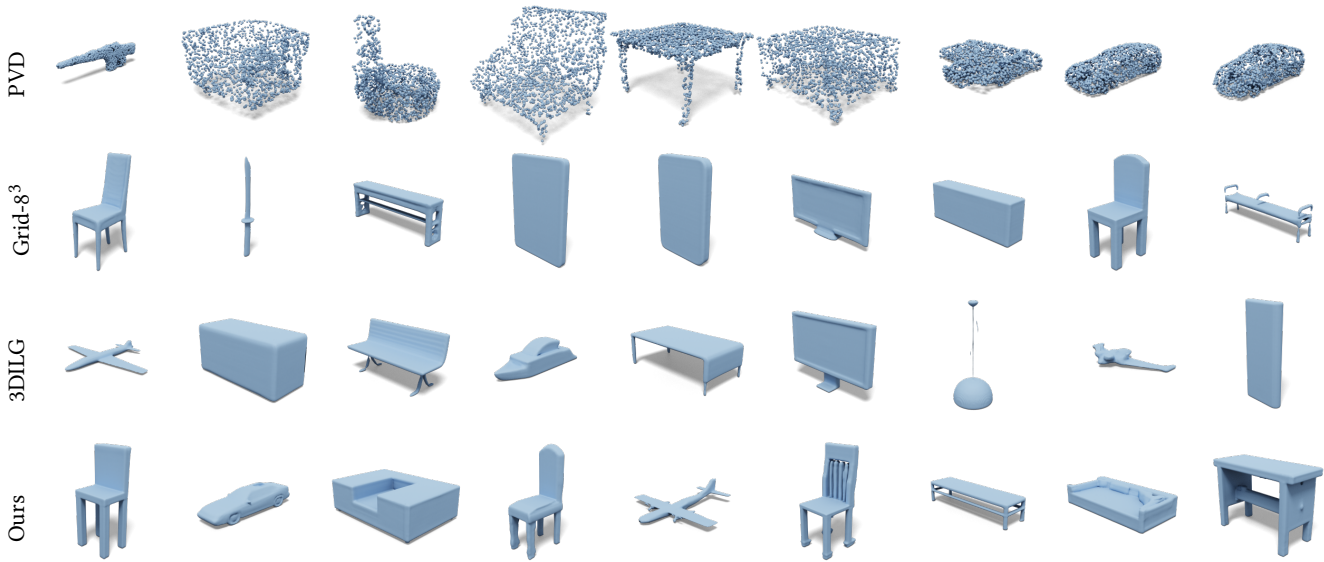
\begin{figure*}[tb]
    \centering
    \begin{overpic}[trim={0cm 0cm 0cm 0cm},clip,width=1\linewidth,grid=false]{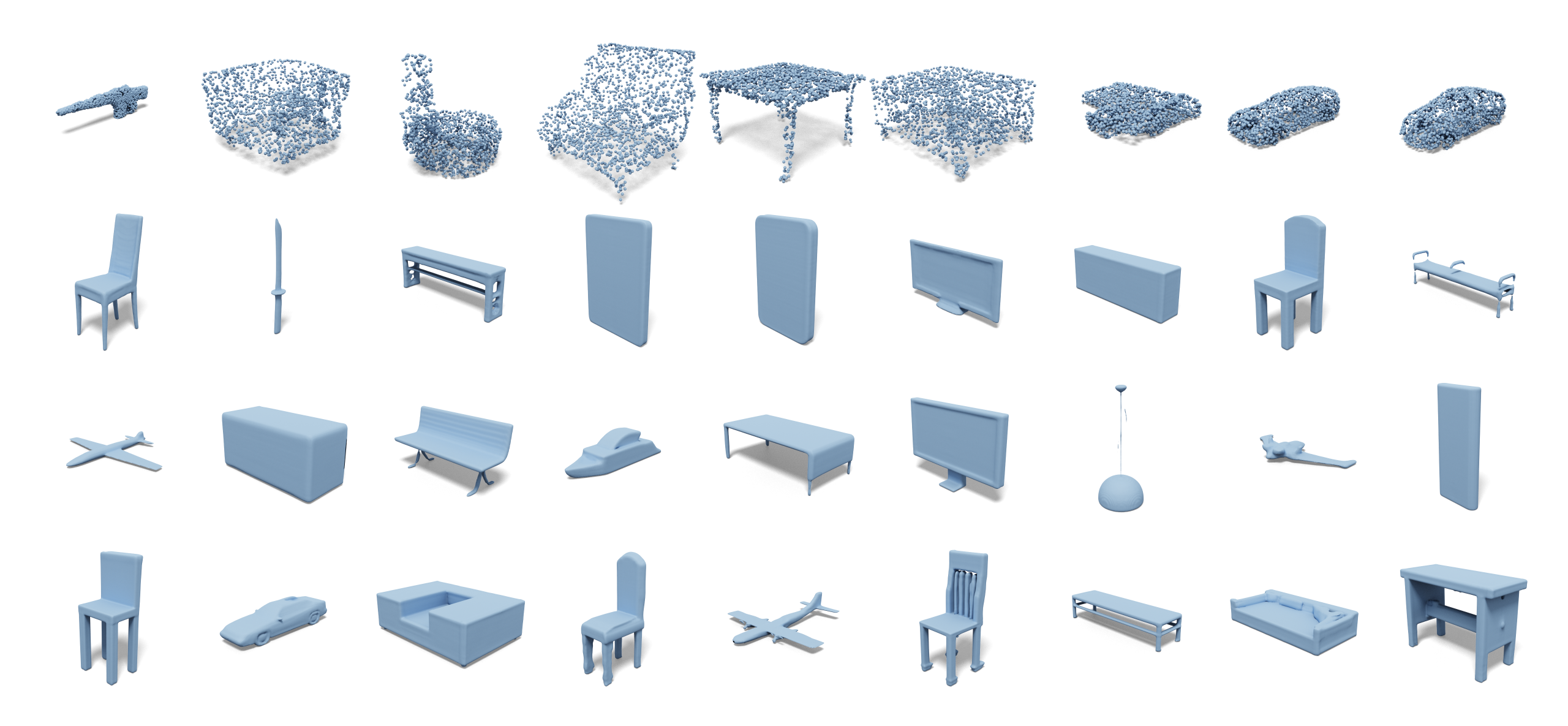}
        \put(0, 5){\rotatebox[]{90}{Ours}}
        \put(0, 16){\rotatebox[]{90}{3DILG}}
        \put(0, 27){\rotatebox[]{90}{Grid-$8^3$}}
        \put(0, 38){\rotatebox[]{90}{PVD}}
    \end{overpic}
    \vspace{-30pt}
    \caption{\textbf{Unconditional generation.} All models are trained on full ShapeNet.}
    \label{fig:uncond-comp}
\end{figure*}

\begin{figure*}[tb]
    \centering
    \vspace*{20pt}
    \begin{overpic}[trim={0cm 1.5cm 0cm 2.5cm},clip,width=1\linewidth,grid=false]{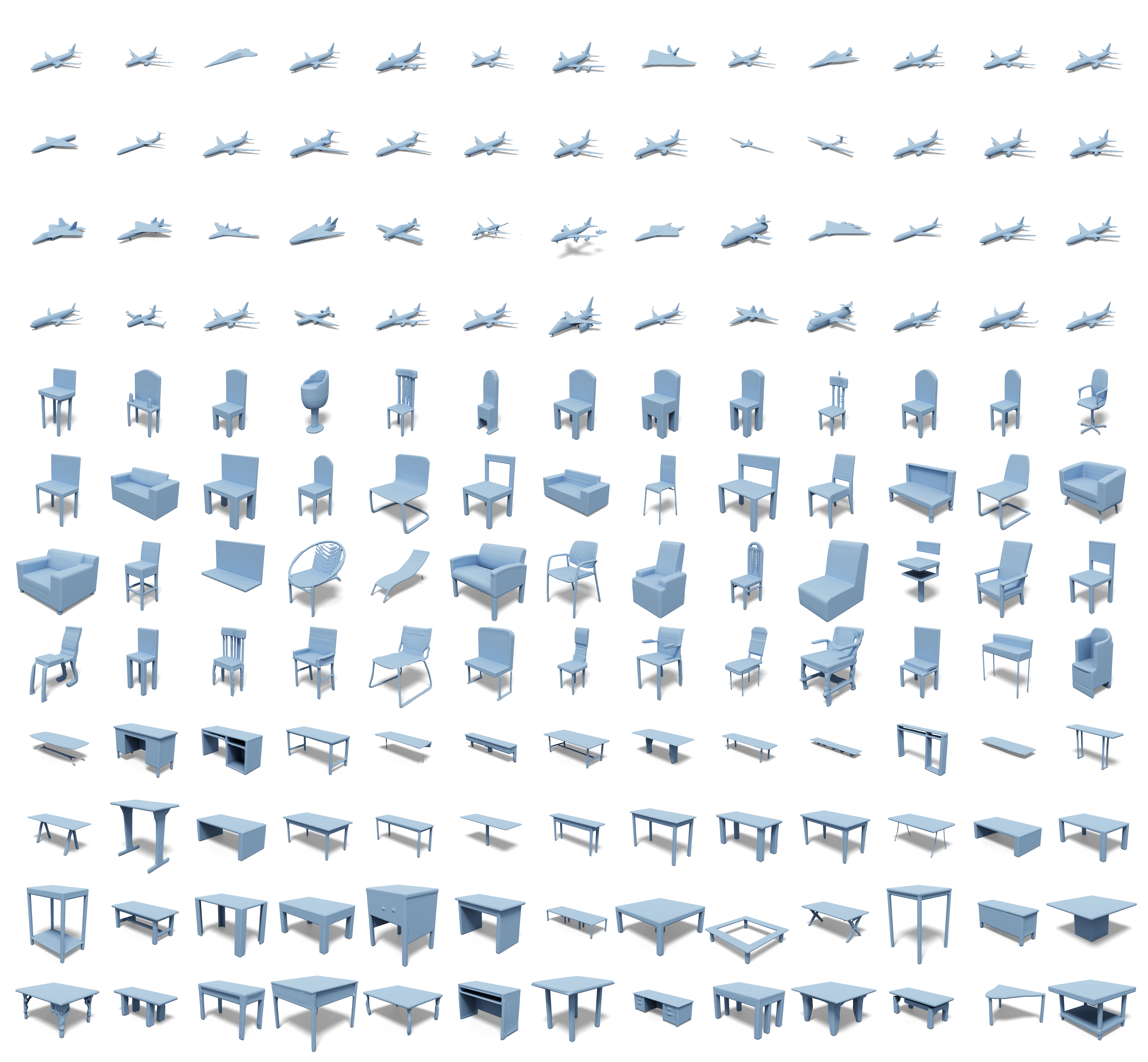}


        
        \dashline{0.7}(0,30)(100,30)
        \dashline{0.7}(0,60)(100,60)
        \put(0, 4){\rotatebox[]{90}{\small{Ours}}}
        \put(0, 11){\rotatebox[]{90}{\small{NW}}}
        \put(0, 19){\rotatebox[]{90}{\small{3DILG}}}
        \put(0, 26){\rotatebox[]{90}{\small{Grid-$8^3$}}}

        \put(0, 33){\rotatebox[]{90}{\small{Ours}}}
        \put(0, 41){\rotatebox[]{90}{\small{NW}}}
        \put(0, 48){\rotatebox[]{90}{\small{3DILG}}}
        \put(0, 56){\rotatebox[]{90}{\small{Grid-$8^3$}}}

        \put(0, 63){\rotatebox[]{90}{\small{Ours}}}
        \put(0, 71){\rotatebox[]{90}{\small{NW}}}
        \put(0, 78){\rotatebox[]{90}{\small{3DILG}}}
        \put(0, 86){\rotatebox[]{90}{\small{Grid-$8^3$}}}
    \end{overpic}
    \caption{\textbf{Category-conditional generation.} From top to bottom, we show category (\emph{airplane, chair, table}) conditioned generation results. }
    \label{fig:cate-cond-comp}
\end{figure*}

We use the dataset of ShapeNet-v2~\cite{chang2015shapenet} as a benchmark, containing 55 categories of man-made objects. We use the training/val splits in~\cite{zhang2022dilg}. We preprocess shapes as in~\cite{mescheder2019occupancy}. Each shape is first converted to a watertight mesh, and then normalized to its bounding box, from which we further sample a dense surface point cloud of size 500,000. To learn the neural fields, we randomly sample 500,000 points with occupancies in the 3D space, and 500,000 points with occupancies in the near surface region.
For the single-view object reconstruction,  we use the 2D rendering dataset provided by 3D-R2N2~\cite{choy20163d}, where each shape is rendered into RGB images of size of $224 \times 224$ from 24 random viewpoints. 
For text-driven shape generation, we use the text prompts of ShapeGlot~\cite{achlioptas2019shapeglot}.
For data preprocess of shape completion training, we create partial point clouds by sampling point cloud patches.

\subsection{Baselines}
For shape auto-encoding, we conduct experiments against state-of-the-art methods for implicit surface reconstruction from point clouds. We use OccNet~\cite{mescheder2019occupancy}, ConvOccNet~\cite{peng2020convolutional}, IF-Net~\cite{chibane2020implicit}, and 3DILG~\cite{zhang2022dilg} as baselines. The OccNet is the first work of learning neural fields from a single global latent vector. ConvOccNet and IF-Net learn local neural fields based on latent vectors arranged in a regular grid, while 3DILG uses latent vectors on an irregular grid.

For 3D shape generation, we compare against recent state-of-the-art generative models, including PVD~\cite{zhou20213d}, 3DILG~\cite{zhang2022dilg}, and NeuralWavelet~\cite{hui2022neural}. PVD is a diffusion model for 3D point cloud generation, and 3DILG utilizes autoregressive models. NeuralWavelet utilized diffusion models in the frequency domain of shapes.
\subsection{Evaluation metrics}
To evaluate the reconstruction accuracy of shape auto-encoding from point clouds, we adopt Chamfer distance, volumetric Intersection-over-Union (IoU), and F-score as primary evaluation metrics. 
IoU is computed based on the occupancy predictions of $50k$ querying points sampled in 3D space.
Chamfer distance and F-score are calculated between two sampled point clouds with the size of $50k$ respectively from reconstructed and ground-truth surfaces. For IoU and F-score, higher is better, while for Chamfer, lower is better. 


To measure the mesh quality of unconditional and conditional shape generation, we follow~\cite{ibing20213d,zhang2022dilg,shue20223d} to adapt the Fr\'echet Inception Distance (FID) and Kernel Inception Distance (KID) commonly used to assess the image generative models to rendered images of 3d shapes. 
To calculate FID and KID of rendered images, we render each shape from 10 viewpoints. The metrics are named as \textbf{Rendering-FID} and \textbf{Rendering-KID}.

The Rendering-FID is defined as,
\begin{equation}
\label{eq:EquaShadingFID}
    \mathrm{Rendering\text{-}FID} = 
    \| \mathbf{\mu_g} - \mathbf{\mu_r} \| + Tr \left( \mathbf\Sigma_g + \mathbf\Sigma_r - 2(\mathbf\Sigma_g\mathbf\Sigma_r)^{1/2} \right) 
\end{equation}
where $g$ and $r$ denotes the generated and training datasets respectively. $\mathbf\mu$ and $\mathbf\Sigma$ are the statistical mean and covariance matrix of the feature distribution extracted by the Inception network.

The Rendering-KID is defined as,
\begin{equation}
\label{eq:EquaShadingKID}
    \mathrm{Rendering\text{-}KID} = 
    \mathrm{MMD}\left(\frac{1}{|\mathcal{R}|} \sum_{\mathbf{x} \in \mathcal{R}} \max\limits_{\mathbf{y} \in \mathcal{G}} D(\mathbf{x}, \mathbf{y})\right)^2 
\end{equation}
where $D(\mathbf{x}, \mathbf{y})$ is a polynomial kernel function to evaluate the similarity of two samples, $\mathcal{G}$ and $\mathcal{R}$ are feature distributions of generated set and reference set, respectively. The function $\mathrm{MMD}(\cdot)$ is Maximum Mean Discrepancy.
However, the rendering-based FID and KID are essentially designed to understand 3D shapes from 2D images. Thus, they have the inherent issue of not accurately understanding shape compositions in the 3D world. To compensate their drawbacks, we also adapt the FID and KID to 3D shapes directly. For each generated or ground-truth shape, we sample 4096 points (with normals) from the surface mesh and then feed them into a pre-trained PointNet++~\cite{qi2017pointnet++} to extract a global latent vector, representing the global structure of the 3D shape. The PointNet++~ is first pretrained on shape classification on ShapeNet-55. As we use point clouds, we call the FID and KID for 3D shapes as Fr\'echet PointNet++ Distance (FPD) and Kernel PointNet++ Distance (KPD). The two metrics are defined similarly as in \cref{eq:EquaShadingFID} and \cref{eq:EquaShadingKID}, except that the features are extracted from a PointNet++ network.



\subsection{Implementation}
For the shape auto-encoder, we use the point cloud of size 2048 as input. At each iteration, we individually sample 1024 query points from the bounding volume ($[-1, 1]^3$) and the other 1024 points from near surface region for the occupancy values prediction. The shape auto-encoder is  trained on 8 A100, with batch size of 512 for $T=1,600$ epochs. The learning rate is linearly increased to $lr_{\text{max}}=5e-5$ in the first $t_0=80$ epochs, and then gradually decreased using the cosine decay schedule $lr_{\text{max}}*0.5^{1+cos(\frac{t-t_0}{T-t_0})}$ until reaching the minimum value of $1e-6$.
The diffusion models are trained on 4 A100 with batch size of 256 for $T=8,000$ epochs. The learning rate is linearly increased to $lr_{max}=1e-4$ in the first $t_0=800$ epochs, and then gradually decreased using the above mentioned decay schedule until reaching $1e-6$.
We use the default settings for the hyperparameters of EDM~\cite{Karras2022edm}. During sampling, we obtain the final latent set via only 18 denoising steps.

\section{Results}
We present our results for multiple applications: 1) shape auto-encoding, 2) unconditional generation, 3) category-conditioned generation, 4) text-conditioned generation, 5) shape completion, 6) image-conditioned generation. Finally, we perform a shape novelty analysis to validate that we are not overfitting to the dataset.

\subsection{Shape Auto-Encoding}\label{sec:result-auto-enc}
We show the quantitative results in \cref{table:main} for a deterministic autoencoder without the KL block described in Sec.~\ref{sec:method-kl}. In particular, we show results for the largest 7 categories as well as averaged results over the categories. The two design choices of shape encoding described in~\cref{sec:method-shape-enc} are also investigated. The case of using the subsampled point cloud as queries is better than learnable queries in all categories. Thus we use subsampled point clouds in our later experiments. The visualization of reconstruction results can be found in~\cref{fig:main}. We visualize some extremely difficult shapes from the datasets (test split). These shapes often contain some thin structures. However, our method still performs well.

Both our method and the competitor 3DILG use transformer as the main backbone. However, we differ in nature. 1) For encoding, 3DILG uses KNN to aggregate local information and we use cross attention. KNN manually selects neighboring points according to spatial similarities (distances) while cross attention learns the similarities on the go. 2) 3DILG uses a set of points and one latent per point. Our representation only contains a set of latents. This simplification makes the second-stage generative model training easier.
3) For decoding, 3DILG applies spatial interpolation and we use interpolation in feature space. The used cross attention can be seen as learnable interpolation. This gives us more flexibility.

The numerical results for the reconstruction are significant. The maximum achievable number for the metrics IoU and F1 is 1. The improvement has to be interpreted in how much closer we get to 1. The visualizations also highlight the improvement.

\paragraph{Ablation study of the number of latents.}
The number $M$ is the number of latent vectors used in the network. Intuitively, a larger $M$ leads to a better reconstruction. We show results of $M$ in~\cref{table:abl-m}. Thus, in all of our experiments, $M$ is set to $512$. We are limited by computation time to work with larger $M$.

\paragraph{Ablation study of the KL block.}
We described the KL block in~\cref{sec:method-kl} that leads to additional compression. In addition, this block changes the deterministic shape encoding into a variational autoencoder. The introduced hyperparameter is $C_0$. A smaller $C_0$ leads to a higher compression rate. The choice of $C_0$ is ablated in~\cref{table:abl-c0}. Clearly, larger $C_0$ gives better results. The reconstruction results of $C_0=8, 16, 32, 64$ are very close. However, they differ significantly in the second stage, because a larger latent size could make the training of diffusion models more difficult. This result is very encouraging for our model, because it indicates that aggressively increasing the compression in the KL block does not decrease reconstruction performance too much.  We can also see that compressing with the KL block by decreasing $C_0$ is much better than compressing using fewer latent vectors $M$.
\subsection{Unconditional Shape Generation}

\paragraph{Comparison with surface generation.}
We evaluate the task of unconditional shape generation with the proposed metrics in~\cref{table:gen-uncond}. We also compared our method with a baseline method proposed in~\cite{zhang2022dilg}. The method is called Grid-$8^3$ because the latent grid size is $8^3$, which is exactly the same as in AutoSDF~\cite{mittal2022autosdf}. The table also shows the results of different $C_0$. Our results are best when $C_0=32$ in all metrics. When $C_0=64$ the results become worse. This also aligns with our conjecture that a larger latent size makes the training more difficult.

\begin{table*}[!htb]
    \begin{minipage}{.4\linewidth}
        \caption{\textbf{Ablation study} for different number of latents $M$ for shape autoencoding}
        \vspace{-10pt}
        \def\arraystretch{1.15}\tabcolsep=0.32em
\begin{tabular}{ccccc}
\toprule
 & $M=512$   & $M=256$   & $M=128$   & $M=64$    \\ \hline
\rowcolor{mytbcol!30} IoU $\uparrow$ & \textbf{0.965} & 0.956 & 0.940 & 0.916 \\
Chamfer $\downarrow$ & \textbf{0.038} & 0.039 & 0.043 & 0.049 \\
\rowcolor{mytbcol!30} F-Score $\uparrow$ & \textbf{0.970} & 0.965 & 0.953 & 0.929 \\ \bottomrule
\end{tabular}

        \label{table:abl-m}
    \end{minipage}\quad
    \begin{minipage}{.55\linewidth}
      \centering
        \caption{
        \textbf{Ablation study} for different number of channels $C_0$ for shape (variational) autoencoding.}
        \vspace{-10pt}

\def\arraystretch{1.15}\tabcolsep=0.32em
\begin{tabular}{cccccccc}
\toprule
 & $C_0=1$     & $C_0=2$     & $C_0=4$              & $C_0=8$              & $C_0=16$             & $C_0=32$     &$C_0=64$         \\ \hline
\rowcolor{mytbcol!30} IoU $\uparrow$ & 0.727 & 0.816 & 0.957          & 0.960          & 0.962          & 0.963 & \textbf{0.964} \\
Chamfer $\downarrow$ & 0.133 & 0.087 & \textbf{0.038} & \textbf{0.038} & \textbf{0.038} & \textbf{0.038} & \textbf{0.038} \\
\rowcolor{mytbcol!30} F-Score $\uparrow$ & 0.703 & 0.815 & 0.967          & 0.967          & \textbf{0.970} & 0.969  & \textbf{0.970}         \\ 
\bottomrule
\end{tabular}

        \label{table:abl-c0}
    \end{minipage} 
\end{table*}

\paragraph{Comparison with point cloud generation.}
Additionally, we compare our method with PVD~\cite{zhou20213d} which is a point cloud diffusion method. We re-train PVD using the official released code on our preprocessed dataset and splits. We use the same evaluation protocol as before but with one major difference. Since PVD can only generate point clouds without normals, we use another pretrained PointNet++ (without normals) as the feature extractor to calculate Surface-FPD and Surface-KPD. The~\cref{table:gen-uncond-pvd} shows we can beat PVD by a large margin. Additionally, we also show the metrics calculated on rendered images. Visualization of generated results can be found in~\cref{fig:uncond-comp}.




\begin{table*}[!htb]
    \begin{minipage}{.6\linewidth}
        \caption{\textbf{Unconditional generation} on full ShapeNet.}
        \vspace{-10pt}
        \def\arraystretch{1.15}
\tabcolsep=0.42em
\begin{tabular}{lcccccc}
\toprule
& \multirow{2}{*}{Grid-$8^3$} & \multirow{2}{*}{3DILG} & \multicolumn{4}{c}{Ours} \\
& & & $C_0=8$ & $C_0=16$ & $C_0=32$ & $C_0=64$\\ \hline
\rowcolor{mytbcol!30}Surface-FPD $\downarrow$ & 4.03 &  1.89 & 2.71  & 1.87  & \textbf{0.76} & 0.97\\
Surface-KPD ($\times 10^3$) $\downarrow$ & 6.15 & 2.17 & 3.48  & 2.42  & \textbf{0.66} & 1.11\\ 
\rowcolor{mytbcol!30} Rendering-FID $\downarrow$ &	32.78 & 24.83	& 28.25	& 27.26	& \textbf{17.08} & 24.24\\
Rendering-KID ($\times 10^3$) $\downarrow$ & 14.12 & 10.51 & 14.60	& 19.37	& \textbf{6.75} & 11.76\\
\bottomrule
\end{tabular}
        \label{table:gen-uncond}
    \end{minipage}%
    \begin{minipage}{.4\linewidth}
      \centering
        \caption{\textbf{Unconditional generation} on full ShapeNet.}
        \vspace{-10pt}
        \def\arraystretch{1.15}
\begin{tabular}{lcc}
\toprule
& \multirow{2}{*}{PVD} & \multirow{2}{*}{Ours}\\
& & \\ \hline
\rowcolor{mytbcol!30}Surface-FPD $\downarrow$ & 2.33 &  \textbf{0.63}\\
Surface-KPD ($\times 10^3$) $\downarrow$ & 2.65 & \textbf{0.53}\\
\rowcolor{mytbcol!30} Rendering-FID $\downarrow$ &	270.64 & \textbf{17.08} \\
Rendering-KID ($\times 10^3$) $\downarrow$ & 281.54 & \textbf{6.75}\\
\bottomrule
\end{tabular}
        \label{table:gen-uncond-pvd}
    \end{minipage} 
\end{table*}

\subsection{Category-conditioned generation}
We train a category-conditioned generation model using our method. We evaluate our models in~\cref{table:class-cond-gen}. We should note that the competitor method NeuralWavelet~\cite{hui2022neural} trains models for categories separately; thus, NeuralWavelet is not a true category-conditioned model. We also visualize some results (\emph{airplane, chair, and table}) in~\cref{fig:cate-cond-comp}. Our training is more challenging, as we train on a dataset that is an order of magnitude larger and we train for all classes jointly. While NeuralWavelet already has good results, the joint training is necessary / beneficial for many subsequent applications.

Additionally, we show evaluation metrics and more competitor methods in~\cref{table:class-cond-gen-detail}. First, we use precision and recall (P\&R)~\cite{sajjadi2018assessing} to quantify the percentage of generated samples that are similar to training and the percentage of training data that can be generated, respectively. 3DILG, NeuralWavelet, and our method, can achieve high precision which means they can generate similar shapes to training. However, our method also shows significantly better recall, which means our method can generate a higher percentage of the training data. For 3DShapeGen and AutoSDF, both precision and recall are low compared to other methods. Second, we show other metrics based on point cloud distances (CD and EMD)~\cite{achlioptas2018learning}. The smaller the better for MMD and the larger the better for COV. These metrics are often used to evaluate point cloud generation.

\begin{table*}[tb]
\centering
\caption{\textbf{Category conditioned generation.} \emph{NW} is short for NeuralWavelet. The dash sign ``-'' means the method NeuralWavelet does not release models trained on these categories.}
\vspace{-10pt}
\def\arraystretch{1.15}\tabcolsep=0.42em
\begin{tabular}{lccc||ccc||ccc||ccc||ccc}

\toprule
& \multicolumn{3}{c||}{airplane} & \multicolumn{3}{c||}{chair} & \multicolumn{3}{c||}{table}& \multicolumn{3}{c||}{car} & \multicolumn{3}{c}{sofa} \\ 
 & 3DILG&  NW   &  Ours            &   3DILG&  NW   &  Ours         &    3DILG&  NW   &  Ours          &    3DILG&  NW   &  Ours         &    3DILG&  NW   &  Ours            \\ \hline
\rowcolor{mytbcol!30}Surface-FID & 0.71 & \textbf{0.38} & 0.62 & 0.96 & 1.14 & \textbf{0.76} & 2.10 & \textbf{1.12} & 1.19 & 2.93 & - & \textbf{2.04} & 1.83 & - & \textbf{0.77} \\
Surface-KID ($\times 10^3$) & 0.81 & \textbf{0.53} & 0.83 & 1.21 & 1.50 & \textbf{0.70} & 3.84 & \textbf{1.55} & 1.87 & 7.35 & - & \textbf{3.90} & 3.36 &  - & \textbf{0.70} \\
 \bottomrule
\end{tabular}

\label{table:class-cond-gen}
\end{table*}

\begin{table*}[]
    \centering
    \caption{\textbf{Category conditioned generation II.} We show results for additional metrics and additional methods for category conditioned generation.}
    \vspace{-10pt}
    \begin{tabular}{cccccc||ccccc}
\toprule
 & \multicolumn{5}{c||}{chair} & \multicolumn{5}{c}{table} \\
 & 3DILG & 3DShapeGen & AutoSDF & NW & Ours &  3DILG & 3DShapeGen & AutoSDF & NW & Ours \\ \hline
\rowcolor{mytbcol!30}Precision $\uparrow$ & 0.87          & 0.56  & 0.42  & \textbf{0.89} & 0.86 & \textbf{0.85} & 0.64  & 0.64  & 0.83  & 0.83           \\
\cellcolor{mytbcol!30}Recall $\uparrow$ & 0.65          & 0.45  & 0.23  & 0.57          & \textbf{0.86} & 0.59          & 0.52  & 0.69  & 0.68  & \textbf{0.89}\\ \hline
\rowcolor{gray!30}MMD-CD $(\times 10^2)$ $\downarrow$ & \textbf{1.78} & 2.14  & 7.27  & 2.14          & \textbf{1.78} & 2.85          & 2.65  & 2.77  & 2.68  & \textbf{2.38}\\
\cellcolor{gray!30}MMD-EMD $(\times 10^2)$ $\downarrow$ & 9.43          & 10.55 & 19.57 & 11.15         & \textbf{9.41} & 11.02         & 9.53  & 9.63  & 9.60  & \textbf{8.81}\\
\rowcolor{gray!30}COV-CD $(\times 10^2)$ $\uparrow$ & 31.95         & 28.01 & 6.31  & 29.19         & \textbf{37.48} & 18.54          & 23.61 & 21.55 & 21.71 & \textbf{25.83}\\
\cellcolor{gray!30}COV-EMD $(\times 10^2)$ $\uparrow$ & 36.29         & 36.69 & 18.34 & 34.91         & \textbf{45.36} & 27.73         & 43.26 & 29.16 & 30.74 & \textbf{43.58}\\ \bottomrule
\end{tabular}

    \label{table:class-cond-gen-detail}
\end{table*}
\subsection{Text-conditioned generation}
\begin{figure}[]
    \centering
    \begin{overpic}[trim={0cm 0cm 0cm 0cm},clip,width=1\linewidth,grid=false]{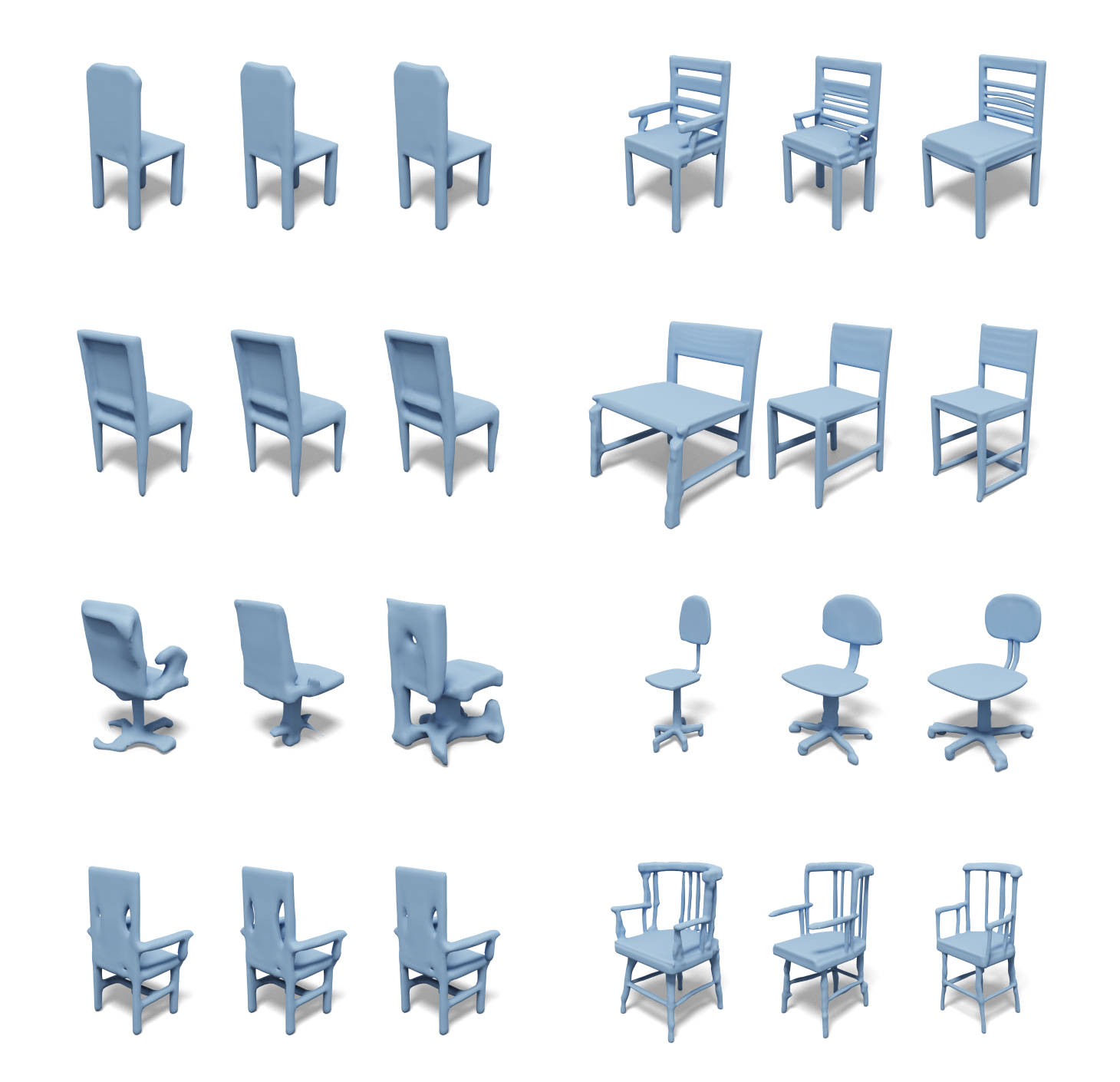}
        \put(21, 93){\small{AutoSDF}}
        \put(70, 93){
        \small{Ours}}
        \put(27, 74){\small{``horizontal slats on top of back''}}
        \put(24, 48){\small{``one big hole between back and seat''}}
        \put(34, 26){\small{``this chair has wheels''}}
        \put(37, 1){\small{``vertical back ribs''}}
    \end{overpic}
    \vspace{-20pt}
    \caption{\textbf{Text conditioned generation.} For each text prompt, we generate 3 shapes. Our results (\textbf{Right}) are compared with AutoSDF (\textbf{Left}).}
    \label{fig:text-cond}
\end{figure}
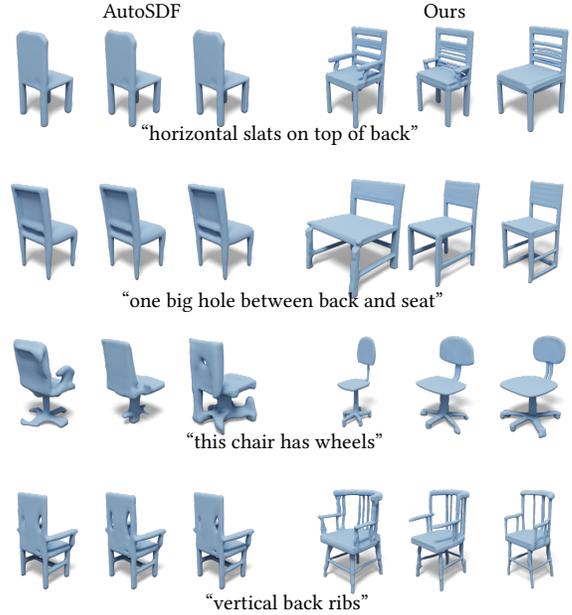

The results of our text-conditioned generation model can be found in~\cref{fig:text-cond}. Since the model is a probabilistic model, we can sample shapes given a text prompt. The results are very encouraging and they constitute the first demonstration of text-conditioned 3D shape generation using diffusion models. To the best of our knowledge, there are no published competing methods at the point of submitting this work.

\subsection{Probabilistic shape completion}
\begin{figure}[]
    \centering
    \begin{overpic}[trim={2cm 1cm 2cm 0cm},clip,width=1\linewidth,grid=false]{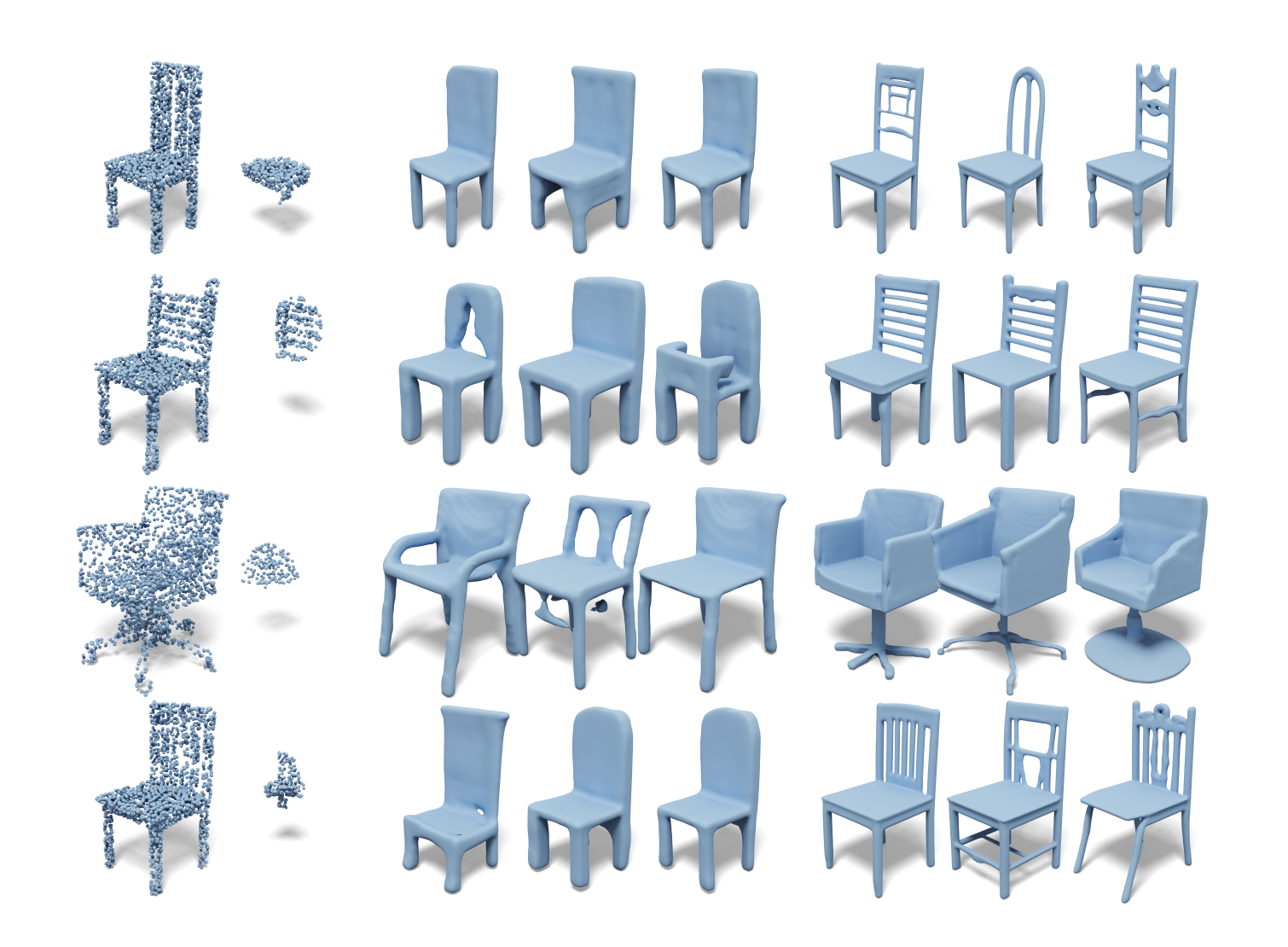}
        \put(7, 73){\small{GT}}
        \put(14, 73){\small{Condition}}
        \put(38, 73){\small{ShapeFormer}}
        \put(77, 73){\small{Ours}}
    \end{overpic}
    \vspace{-20pt}
    \caption{\textbf{Point cloud conditioned generation.} We show three generated results given a partial cloud.
    The ground-truth point cloud and the partial point cloud used as condition are shown in \textbf{Left}. We compare our results (\textbf{Right}) with ShapeFormer (\textbf{Middle}).}
    \label{fig:point-cond}
\end{figure}

We also extend our diffusion model for probabilistic shape completion by using a partial point cloud as conditioning input. The comparison against ShapeFormer~\cite{yan2022shapeformer} is depicted in~\cref{fig:point-cond}. As seen, our latent set diffusion can predict more accurate completion, and we also have the ability to achieve more diverse generations.

\subsection{Image-conditioned shape generation.} 
\begin{figure}[]
    \centering
    \begin{overpic}[trim={2cm 1cm 2cm 0cm},clip,width=1\linewidth,grid=false]{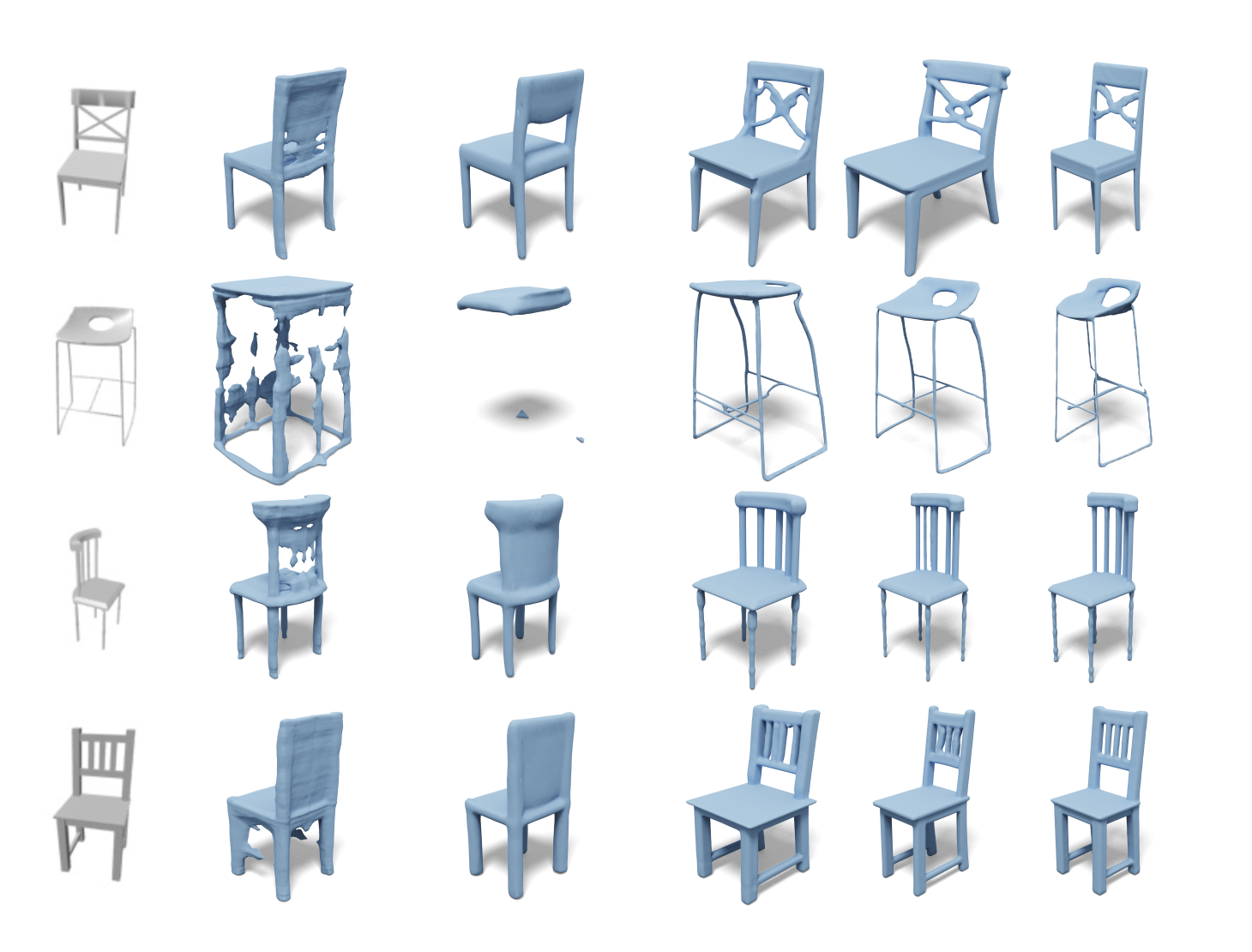}
        \put(-1, 77){\small{Condition}}
        \put(17, 77){\small{IM-Net}}
        \put(36, 77){\small{OccNet}}
        \put(72, 77){\small{Ours}}
    \end{overpic}
    \vspace{-20pt}
    \caption{\textbf{Image conditioned generation.} In the \textbf{left} column we show the condition image. In the \textbf{middle} we show results obtained by the method IM-Net and OccNet. Our generated results are shown on the \textbf{right}.}
    \label{fig:img-cond}
\end{figure}
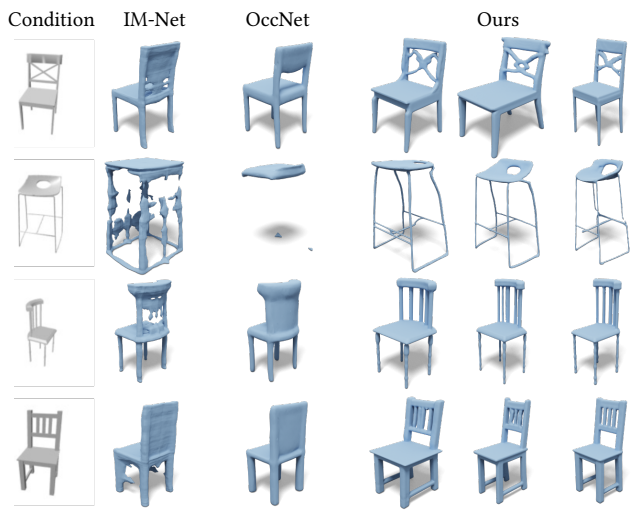

We also provide comparisons on the task of single-view 3D object reconstruction in~\cref{fig:img-cond}. Compared to other deterministic methods including OccNet~\cite{mescheder2019occupancy} and IM-Net~\cite{chen2019learning}, our latent set diffusion can not only reconstruct more accurate surface details, (e.g. long rods and tiny holes in the back), but also support multi-modal prediction, which is a desired property to deal with severe occlusions.

\subsection{Shape novelty analysis}
\begin{figure}[]
    \centering
    \begin{overpic}[trim={2cm 0cm 2cm 0cm},clip,width=1\linewidth,grid=false]{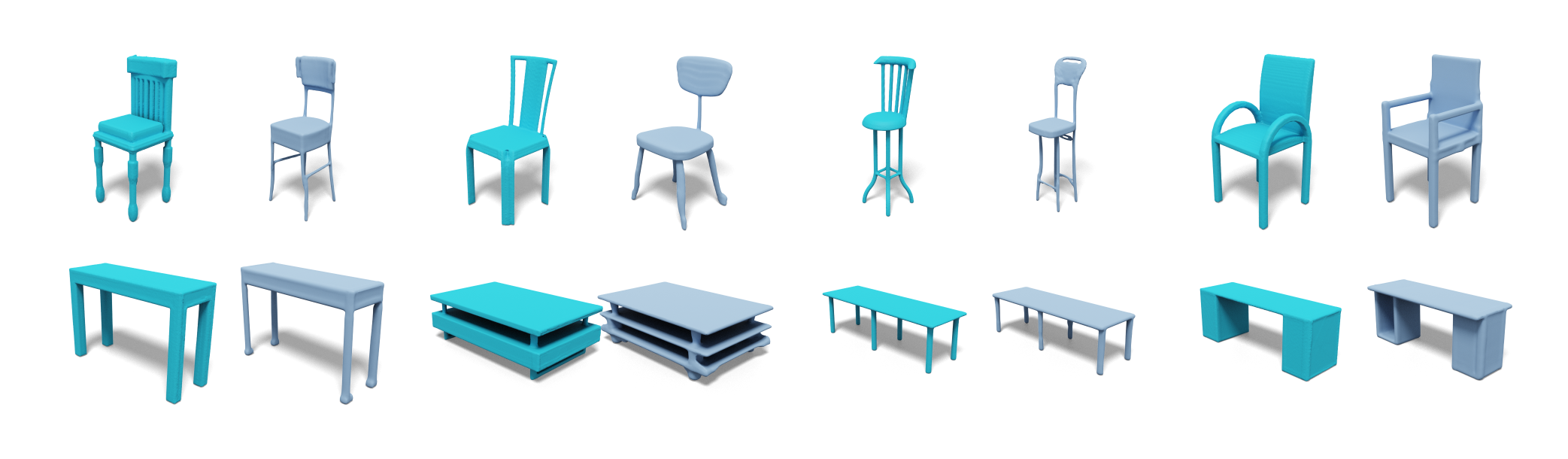}
        \put(3, 29){\small{Ref}}
        \put(14, 29){\small{Gen}}
        \put(29, 29){\small{Ref}}
        \put(41, 29){\small{Gen}}
        \put(55, 29){\small{Ref}}
        \put(67, 29){\small{Gen}}
        \put(80, 29){\small{Ref}}
        \put(91, 29){\small{Gen}}
    \end{overpic}
    \vspace{-20pt}
    \caption{\textbf{Shape generation novelty.} For a generated shape, we retrieve the top-1 similar shape in the training set. The similarity is measured using Chamfer distance of sampled surface point clouds. In each pair, we show the retrieved shape (\textbf{left}) and the generated shape (\textbf{right}). The generated shapes are from our category-conditioned generation results.}
    \label{fig:novelty}
\end{figure}
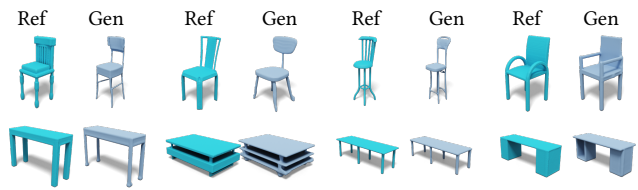

We use shape retrieval to demonstrate that we are not simply overfitting to the training set. Given a generated shape, we measure the Chamfer distance between it and training shapes. The visualization of retrieved shapes can be found in~\cref{fig:novelty}. Clearly, the model can synthesize new shapes with realistic structures.

\subsection{Limitations}
While our method shows convincing results on a variety of tasks, our design choices also have drawbacks that we would like to discuss. 
For instance, we require a two stage training strategy. 
While this leads to improved performance in terms of generation quality, training the first stage is more time consuming than relying on manually-designed features such as wavelets~\cite{hui2022neural}.
In addition, the first stage might require retraining if the shape data in consideration changes, and for the second stage -- the core of our diffusion architecture -- training time is also relatively high.
Overall, we believe that there is significant potential for future research avenues to speed up training, in particular, in the context of diffusion models. 

\section{Conclusion}
We have introduced 3DShape2VecSet, a novel shape representation for neural fields that is tailored to generative diffusion models.
To this end, we combine ideas from radial basis functions, previous neural field architectures, variational autoencoding, as well as cross attention and self-attention to design a learnable representation.
Our shape representation can take a variety of inputs including triangle meshes and point clouds and encode 3D shapes as neural fields on top of a set of latent vectors.
As a result, our method demonstrates improved performance in 3D shape encoding and 3D shape generative modeling tasks, including unconditioned generation, category-conditioned generation, text-conditioned generation, point-cloud completion, and image-conditioned generation.

In future work, we see many exciting possibilities. 
Most importantly, we believe that our model further advances the state of the art in point cloud and shape processing on a large variety of tasks. 
In particular, we would like to employ the network architecture of 3DShape2VecSet to tackle the problem of surface reconstruction from scanned point clouds.
In addition, we can see many applications for content-creation tasks, for example 3D shape generation of textured models along with their material properties.
Finally, we would like to explore editing and manipulation tasks leveraging pretrained diffusion models for prompt to prompt shape editing, leveraging the recent advances in image diffusion models.

\begin{acks}

We would like to acknowledge Anna Frühstück for helping with figures and the video voiceover. This work was supported by the SDAIA-KAUST Center of Excellence in Data Science and Artificial Intelligence (SDAIA-KAUST AI) as well as the ERC Starting Grant
Scan2CAD (804724).
\end{acks}

\bibliographystyle{ACM-Reference-Format}
\bibliography{bib}

\end{document}